\documentclass{sig-alternate-2013}
\usepackage[ruled]{algorithm2e}
\usepackage{subfigure,caption}

\usepackage{hyperref}
\usepackage{url}
\usepackage{multirow,amsmath,mathrsfs,color,verbatim}
\usepackage{graphicx}
\usepackage{color}

\permission{Copyright is held by the International World Wide Web Conference Committee (IW3C2). IW3C2 reserves the right to provide a hyperlink to the author's site if the Material is used in electronic media.}
\conferenceinfo{WWW 2016,}{April 11--15, 2016, Montr\'eal, Qu\'ebec, Canada.} 
\copyrightetc{ACM \the\acmcopyr}
\crdata{978-1-4503-4143-1/16/04. \\
http://dx.doi.org/10.1145/2872427.2883041 }

\begin{document}
%

\title{Visualizing Large-scale and High-dimensional Data}

\author{
	Jian Tang$^1$, Jingzhou Liu$^2$\thanks{This work was done when the second author was an intern at Microsoft Research Asia.}, Ming Zhang$^{2}$, Qiaozhu Mei$^3$ \\
	\affaddr{$^1$Microsoft Research Asia, jiatang@microsoft.com}\\
	\affaddr{$^2$Peking University, \{liujingzhou, mzhang\_cs\}@pku.edu.cn }\\
		\affaddr{$^3$University of Michigan,  qmei@umich.edu}\\
}

\maketitle
\begin{abstract}
We study the problem of visualizing large-scale and high-dimensional data in a low-dimensional (typically 2D or 3D) space. Much success has been reported recently by techniques that first compute a similarity structure of the data points and then project them into a low-dimensional space with the structure preserved. These two steps suffer from considerable computational costs, preventing the state-of-the-art methods such as the t-SNE from scaling to large-scale and high-dimensional data (e.g., millions of data points and hundreds of dimensions). We propose the LargeVis, a technique that first constructs an accurately approximated K-nearest neighbor graph from the data and then layouts the graph in the low-dimensional space. Comparing to t-SNE, LargeVis significantly reduces the computational cost of the graph construction step and employs a principled probabilistic model for the visualization step, the objective of which can be effectively optimized through asynchronous stochastic gradient descent with a linear time complexity. The whole procedure thus easily scales to millions of high-dimensional data points. Experimental results on real-world data sets demonstrate that the LargeVis outperforms the state-of-the-art methods in both efficiency and effectiveness. The hyper-parameters of LargeVis are also much more stable over different data sets. 

\end{abstract}

\terms{Algorithms, Experimentation}

\section{Introduction}
\label{sec::intro}

We now live in the era of the big data. Understanding and mining large-scale data sets have created big opportunities for business providers, data scientists, governments, educators, and healthcare practitioners. Many computational infrastructures, algorithms, and tools are being constructed for the users to manage, explore, and analyze their data sets. Information visualization has been playing a critical role in this pipeline, which facilitates the description, exploration, and sense-making from both the original data and the analysis results \cite{keim2002information}. Classical visualization techniques have been proved effective for small or intermediate size data; they however face a big challenge when applied to the big data. For example, visualizations such as scatter plots, heatmaps, and network diagrams all require laying out data points on a 2D or 3D space, which becomes computationally intractable when there are too many data points and when the data have many dimensions. Indeed, while there exist numerous network diagrams with thousands of nodes, a visualization of millions of nodes is rare, even if such a visualization would easily reveal node centrality and community structures. In general, the problem is concerned with finding an extremely low-dimensional (e.g., 2D or 3D) representation of large-scale and high-dimensional data, which has attracted a lot of attentions recently in both the data mining community \cite{torgerson1952multidimensional,belkin2001laplacian,van2008visualizing} and the infoviz community \cite{fruchterman1991graph,jacomy2011forceatlas2,martin2011openord}.  Compared to the high-dimensional representations, the 2D or 3D layouts not only demonstrate the intrinsic structure of the data intuitively and can also be used as the basis to build many advanced and interactive visualizations. 

\begin{figure*}[htdb!]
	\centering
	\includegraphics[width=0.75\textwidth]{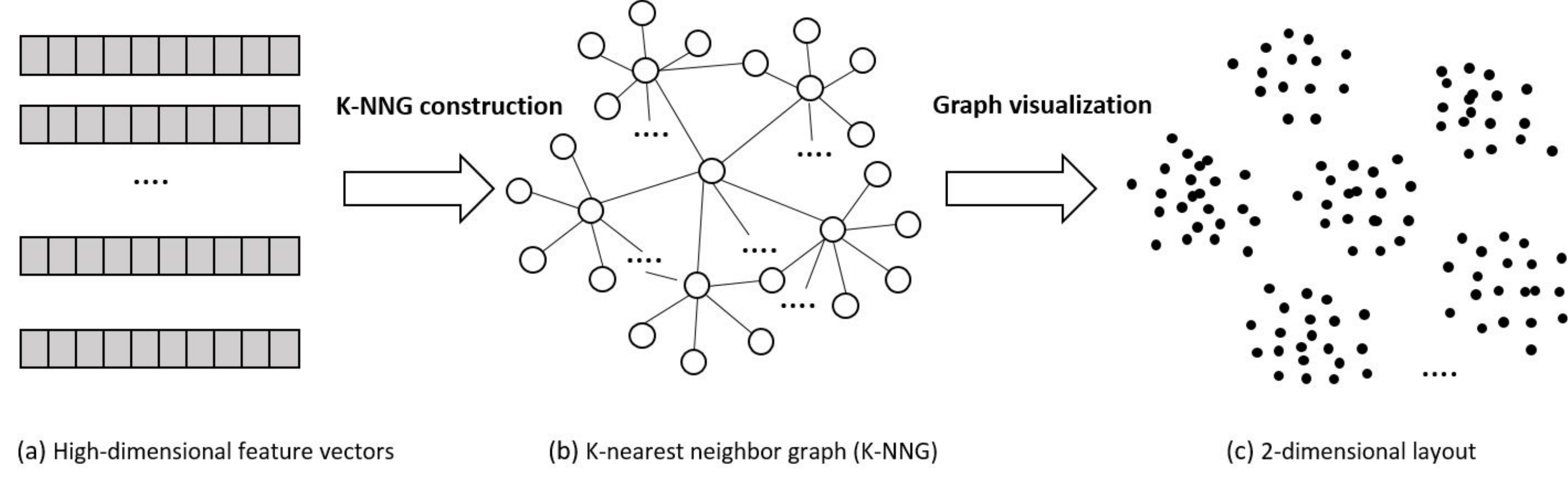}
	\caption{A typical pipeline of data visualization by first constructing a K-nearest neighbor graph and then projecting the graph into a low-dimensional space.}
	\label{fig::vis_pipeline}
\end{figure*}


Projecting high-dimensional data into spaces with fewer dimensions is a core problem of machine learning and data mining. The essential idea is to preserve the intrinsic structure of the high-dimensional data, i.e., keeping similar data points close and dissimilar data points far apart, in the low-dimensional space. In literature, many dimensionality reduction techniques have been proposed, including both linear mapping methods (e.g., Principle Component Analysis~\cite{jolliffe2002principal}, multidimensional scaling~\cite{torgerson1952multidimensional}) and non-linear mapping methods (e.g., Isomap~\cite{tenenbaum2000global}, Locally Linear Embedding~\cite{roweis2000nonlinear}, Laplacian Eigenmaps~\cite{belkin2001laplacian}). As most high-dimensional data usually lie on or near a low-dimensional non-linear manifold, the performance of linear mapping approaches is usually not satisfactory~\cite{van2008visualizing}. For non-linear methods such as the Laplacian Eigenmaps, although empirically effective on small, laboratory data sets, they generally do not perform well on high-dimensional, real data as they are typically not able to preserve both the local and the global structures of the high-dimensional data. Maaten and Hinton proposed the t-SNE technique~\cite{van2008visualizing}, which captures both the local and the global structures.  Maaten further proposed an acceleration technique~\cite{van2014accelerating} for the t-SNE by first constructing a K-nearest neighbor (KNN) graph of the data points and then projecting the graph into low-dimensional spaces with tree-based algorithms. T-SNE and its variants, which represent a family of methods that first construct a similarity structure from the data and then project the structure into a 2D/3D space (see Figure~\ref{fig::vis_pipeline}), have been widely adopted recently due to the ability to handle real-world data and the good quality of visualizations.

Despite their successes, when applied to data with millions of points and hundreds of dimensions, the t-SNE techniques are still far from satisfaction. The reasons are three-fold: (1) the construction of the K-nearest neighbor graph is a computational bottleneck for dealing with large-scale and high-dimensional data. T-SNE constructs the graph using the technique of vantage-point trees~\cite{yianilos1993data}, the performance of which significantly deteriorates when the dimensionality of the data grows high; (2) the efficiency of the graph visualization step significantly deteriorates when the size of the data becomes large; (3) the parameters of the t-SNE are very sensitive on different data sets. To generate a good visualization, one has to search for the optimal parameters exhaustively, which is very time consuming on large data sets. It is still a long shot of the community to create high quality visualizations that scales to both the size and the dimensionality of the data. 

We report a significant progress on this direction through the LargeVis, a new visualization technique that computes the layout of large-scale and high-dimensional data. The LargeVis employs a very efficient algorithm to construct an approximate K-nearest neighbor graph at a high accuracy, which builds on top of but significantly improves a state-of-the-art approach to KNN graph construction, the random projection trees~\cite{dasgupta2008random}. 
We then propose a principled probabilistic approach to visualizing the K-nearest neighbor graph, which models both the observed links and the unobserved (i.e., negative) links in the graph. The model preserves the structures of the graph in the low-dimensional space, keeping similar data points close and dissimilar data points far away from each other. The corresponding objective function can be optimized through the asynchronous stochastic gradient descent, which scales \textit{linearly} to the data size $N$. Comparing to the one used by the t-SNE, the optimization process of LargeVis is much more efficient and also more effective. Besides, on different data sets the parameters of the LargeVis are much more stable.

We conduct extensive experiments on real-world, large-scale and high-dimensional data sets, including text (words and documents), images, and networks. Experimental results show that our proposed algorithm for constructing the approximate K-nearest neighbor graph significantly outperforms the vantage-point tree algorithm used in the t-SNE and other state-of-the-art methods. 
LargeVis generates comparable graph visualizations to the t-SNE on small data sets and more intuitive visualizations on large data sets; it is much more efficient when data becomes large; the parameters are not sensitive to different data sets. On a set of three million data points with one hundred dimensions, LargeVis is up to thirty times faster at graph construction and seven times faster at graph visualization. LargeVis only takes a couple of hours to visualize millions of data points with hundreds of dimensions on a single machine. 

To summarize, we make the following contributions:

\begin{itemize}
	\item We propose a new visualization technique which is able to compute the layout of millions of data points with hundreds of dimensions efficiently. 
	\item We propose a very efficient algorithm to construct an approximate K-nearest neighbor graph from large-scale, high-dimensional data. 
	\item We propose a principled probabilistic model for graph visualization. The objective function of the model can be effectively optimized through asynchronous stochastic gradient descent with a time complexity of $O(N)$.
	\item We conduct experiments on real-world, very large data sets and compare the performance of LargeVis and t-SNE, both quantitatively and visually. 
\end{itemize}

\section{Related Work}
\label{sec::related}

To the best of our knowledge, very few visualization techniques can efficiently layout millions of high-dimensional data points meaningfully on a 2D space. Instead, most visualizations of large data sets have to first layout a summary or a coarse aggregation of the data and then refine a subset of the data (a region of the visualization) if the user zooms in \cite{card1999readings}. Admittedly, there are other design factors besides the computational capability, for example the aggregated data may be more intuitive and more robust to noises. However, with a layout of the entire data set as basis, the effectiveness of these aggregated/approximate visualizations will only be improved. Many visualization tools are designed to layout geographical data, sensor data, and network data. These tools typically cannot handle high-dimensional data. 


Many recent successes of visualizing high-dimensional data come from the machine learning community. Methods like the t-SNE first compute a K-nearest-neighbor graph and then visualizes this graph in a 2D/3D space. Our work follows this direction and makes significant progress. 

\subsection{K-nearest Neighbor Graph Construction}
Constructing K-nearest neighbor (KNN) graphs from high-dimensional data is critical to many applications such as similarity search, collaborative filtering, manifold learning, and network analysis. While the exact computation of a KNN has a complexity of $O(N^2d)$ (with $N$ being the number of data points and $d$ being the number of dimensions) which is too costly, existing approaches use roughly three categories of techniques: space-partitioning trees~\cite{bentley1975multidimensional,friedman1977algorithm,silpa2008optimised, dasgupta2008random}, locality sensitive hashing techniques~\cite{datar2004locality,charikar2002similarity,gionis1999similarity}, and neighbor exploring techniques~\cite{dong2011efficient}. The space-partitioning methods divide the entire space into different regions and organize the regions into different tree structures, e.g., k-d trees~\cite{bentley1975multidimensional,friedman1977algorithm}, vp-trees~\cite{yianilos1993data}, cover trees~\cite{beygelzimer2006cover}, and random projection trees~\cite{dasgupta2008random}. Once the trees are constructed, the nearest neighbors of each data point can be found through traversing the trees. The locality sensitive hashing~\cite{datar2004locality} techniques deploy multiple hashing functions to map the data points into different buckets and data points in the same buckets are likely to be similar to each other. The neighbor exploring techniques, such as the NN-Descent~\cite{dong2011efficient}, is built on top of the intuition that ``my neighbors' neighbors are likely to be my neighbors.'' Starting from an initial nearest-neighbor graph, the algorithm iteratively refines the graph by exploring the neighbors of neighbors defined according to the current graph.

The above approaches work efficiently on different types of data sets. The k-d trees, vp-trees, or cover-trees have been proved to very efficient on data with a small number of dimensions. However, the performance significantly deteriorates when the dimensionality of the data becomes large (e.g., hundreds). The NN-descent approach is also usually efficient for data sets with a small number of dimensions~\cite{dong2011efficient}. 
A comparison of these techniques can be found at \url{https://github.com/erikbern/ann-benchmarks}. The random projection trees have demonstrated state-of-the-art performance in constructing very accurate K-nearest neighbor graphs from high-dimensional data. However, the high accuracy is at the expense of efficiency, as to achieve a higher accuracy many more trees have to be created. Our proposed technique is built upon random projection trees but significantly improves it using the idea of neighbor exploring. 
The accuracy of a KNN graph quickly improves to almost 100\% without investing in many trees. 
  
\subsection{Graph Visualization}
The problem of graph visualization is related to dimensionality reduction, which includes two major types of approaches: linear transformations and non-linear transformations. When projecting the data to extremely low-dimensional spaces (e.g., 2D), the linear methods such as the Principle Component Analysis~\cite{jolliffe2002principal} and the multidimensional scaling~\cite{torgerson1952multidimensional} usually do not work as effectively as the non-linear methods as most high-dimensional data usually lies on or near low-dimensional non-linear manifolds. The non-linear methods such as Isomap~\cite{tenenbaum2000global}, local linear embedding (LLE)~\cite{roweis2000nonlinear}, Laplacian Eigenmaps~\cite{belkin2001laplacian} are very effective on laboratory data sets but do not perform really well on real-world high-dimensional data. Maaten and Hinton proposed the t-SNE~\cite{van2008visualizing}, which is very effective on real-world data. None of these methods scales to millions of data points. Maaten improved the efficiency of t-SNE through two tree based algorithms ~\cite{van2014accelerating}, which scale better to large graphs. The optimization of the t-SNE requires the fully batch gradient descent learning, the time complexity of which w.r.t the data size $N$ is $O(N\log N)$. LargeVis can be naturally optimized through asynchronous stochastic gradient descent, with a complexity of $O(N)$. Besides, the parameters of t-SNE are very sensitive on different sets while the parameters of LargeVis remain very stable. 

There are many algorithms developed in the information visualization community to compute the layout of nodes in a network. They can also be used to visualize the KNN graph. The majority of these network layout methods use either the abovementioned dimensionality reduction techniques or force-directed simulations. Among them, force-directed layouts generates better visualizations, but their high computational complexity (ranging from $O(N^3)$ to $O(N \log^2 N)$ with $N$ being the number of nodes \cite{tamassia2013handbook}) has prevented them from being applied to millions of nodes.

Among them, the classical Fruchterman-Reingo algorithm \cite{fruchterman1991graph} and the original ForceAtlas algorithm provided in Gephi \cite{bastian2009gephi} have a complexity of $O(N^2)$. An improved version of ForceAtlas called the ForceAtlas2 \cite{jacomy2011forceatlas2} and the newly developed Openord algorithm~\cite{martin2011openord} reduce the time complexity to $O(N\log N)$. These two algorithms have been used to visualize one million data points~\footnote{\url{http://sebastien.pro/gephi-esnam.pdf}}, but the complexity prevents them from scaling up further. 

The LargeVis is also related to our previous work on network/graph embedding, the LINE model~\cite{tang2015line}. LINE and other related methods (e.g., Skipgram \cite{mikolov2013distributed}) are not designed for visualization purposes. Using them directly to learn 2/3-dimensional representations of data may yield ineffective visualization results. However, these methods can be used as a preprocessor of the data for the visualization (e.g., use LINE or Skipgram to learn 100 dimensional representations of the data and then use LargeVis to visualize them). 
\section{LargeVis}
\label{sec::definition}
In this section, we introduce the new visualization technique LargeVis. Formally, given a large-scale and high-dimensional data set $\mathcal{X} = \{\vec{x}_i\in R^d\}_{i=1,2,\ldots, N}$, our goal is to represent each data point $\vec{x}_i$ with a low-dimensional vector $\vec{y}_i\in R^s$, where $s$ is typically 2 or 3. The basic idea of visualizing high-dimensional data is to preserve the intrinsic structure of the data in the low-dimensional space. Existing approaches usually first compute the similarities of all pairs of $\{\vec{x}_i, \vec{x}_j\}$ and then preserve the similarities in the low-dimensional transformation. As computing the pairwise similarities is too expensive (i.e., $O(N^2d)$), recent approaches like the t-SNE construct a K-nearest neighbor graph instead and then project the graph into the 2D space. LargeVis follows this procedure, but uses a very efficient algorithm for K-nearest neighbor graph construction and a principled probabilistic model for graph visualization. 
Next, we introduce the two components respectively. 

\subsection{Efficient KNN Graph Construction}
\label{sec::knn}
A K-nearest neighbor graph requires a metric of distance. We use the Euclidean distance $||\vec{x}_i-\vec{x_j}||$ in the high-dimensional space, the same as the one used by t-SNE. Given a set of high-dimensional data points $\{\vec{x}_i\}_{i=1,\ldots, N}$, in which $\vec{x}_i \in R^d$,  constructing the exact KNN graph takes $O(N^2d)$ time - too costly. Various indexing techniques have been proposed to approximate the KNN graph (see Section~\ref{sec::related}). 

Among these techniques, the random projection trees have been 
proved to be very efficient for nearest-neighbor search in high-dimensional data.  The algorithm starts by partitioning the entire space and building up a tree. Specifically, for every non-leaf node of the tree, the algorithm selects a random hyperplane to split the subspace corresponding to the non-leaf node into two, which become the children of that node. The hyperplane is selected through randomly sampling two points from the current subspace and then taking the hyperplane equally distant to the two points. This process continues until the number of nodes in the subspace reaches a threshold. Once a random projection tree is constructed, every data point can traverse the tree to find a corresponding leaf node. The points in the subspace of that leaf node will be treated as the candidates of the nearest neighbors of the input data point.  In practice multiple trees can be built to improve the accuracy of the nearest neighbors. Once the nearest neighbors of all the data points are found, the K-nearest neighbor graph is built. 

However, constructing a very accurate KNN graph requires many trees to be built, which significantly hurts the efficiency. This dilemma has been a bottleneck of applying random projection trees to visualization. In this paper we propose a new solution: instead of building a large number of trees to obtain a highly accurate KNN graph, we use neighbor exploring techniques to improve the accuracy of a less accurate graph. The basic idea is that ``a neighbor of my neighbor is also likely to be my neighbor''~\cite{dong2011efficient}. Specifically, we build a few random projection trees to construct an approximate K-nearest neighbor graph, the accuracy of which may be not so high. Then for each node of the graph, we search the neighbors of its neighbors, which are also likely to be candidates of its nearest neighbors. We may repeat this for multiple iterations to improve the accuracy of the graph. In practice, we find that only a few iterations are sufficient to improve the accuracy of the KNN graph to almost 100\%. 

For the weights of the edges in the K-nearest neighbor graph, we use the same approach as t-SNE. The conditional probability from data $\vec{x}_i$ to $\vec{x}_j$ is first calculated as:
\begin{equation}
\label{eqn::similarity_1}
	\begin{aligned}
	 p_{j|i}&=\frac{\exp(-||\vec{x}_i-\vec{x}_j||^2/2\sigma_i^2)}{\sum_{(i,k)\in E} \exp(-||\vec{x}_i-\vec{x}_k||^2/2\sigma_i^2)}, ~\mbox{and}    \\
	  p_{i|i}&=0,
	 \end{aligned}
\end{equation}
where the parameter $\sigma_i$ is chosen by setting the perplexity of the conditional distribution $p_{\cdot|i}$ equal to a perplexity $u$.  Then the graph is symmetrized through setting the weight between $\vec{x}_i$ and $\vec{x}_j$ as:
\begin{equation}
\label{eqn::similarity_2}
w_{ij}=\frac{p_{j|i}+p_{i|j}}{2N}.
\end{equation}

The complete procedure is summarized in Algo.~\ref{algo::graph_construction}.

\begin{algorithm}[!htdb]
	\scriptsize
\KwData{ $\{\vec{x}_i\}_{i=1,\ldots, N}$, number of trees $NT$, number of neighbors $K$, number of iterations $Iter$.}
\KwResult{Approximate K-nearest neighbor graph $G$.}
\SetAlgoNoLine
1. Build $NT$ random projection trees on $\{\vec{x}_i\}_{i=1,\ldots, N}$\;
2. Search nearest neighbors: \\
\Indp
\For{each node $i$ in parallel}
{
	Search the random projection trees for $i$'s $K$ nearest neighbors, store the results in $knn(i)$\;
}
\Indm
3. Neighbor exploring: \\
\Indp
\While{$T < Iter$}
{
	Set $old\_knn() = knn()$, clear $knn()$\;
	\For{each node $i$ in parallel}
	{
		Create max heap $H_i$\;
		\For{$j \in old\_knn(i)$}
		{
			\For{$l \in old\_knn(j)$}
			{
				Calculate $dist(i,l)=||\vec{x}_i - \vec{x}_l||$\;
				Push $l$ with $dist(i,l)$ into $H_i$\;
				Pop if $H_i$ has more than $K$ nodes\;
			}
		}
		Put nodes in $H_i$ into $knn(i)$\;
	}
	T++\;
}
\For{each node $i$ and each $j \in knn(i)$}
{
	Add edge $(i,j)$ into graph $G$\;
}
\Indm
4. Calculate the weights of the edges according to Eqn.~\ref{eqn::similarity_1},~\ref{eqn::similarity_2}. 
	\caption{Graph Construction}
	\label{algo::graph_construction}
\end{algorithm}

\subsection{A Probabilistic Model for Graph Visualization}
\label{sec::visualization}
Once the KNN graph is constructed, to visualize the data we just need to project the nodes of the graph into a 2D/3D space. We introduce a principled probabilistic model for this purpose. The idea is to preserve the similarities of the vertices in the low-dimensional space. In other words, we want to keep similar vertices close to each other and dissimilar vertices far apart in the low-dimensional space.
Given a pair of vertices $(v_i, v_j)$, we first define the probability of observing a binary edge $e_{ij}=1$ between $v_i$ and $v_j$ as follows:
\begin{equation}
	\label{eqn::edge_probability}
	P(e_{ij}=1) =f(||\vec{y}_i-\vec{y}_j||),
\end{equation}
where $\vec{y}_i$ is the embedding of vertex $v_i$ in the low-dimensional space, $f(\cdot)$ is a probabilistic function w.r.t the distance of vertex $y_i$ and $y_j$, i.e., $||\vec{y}_i-\vec{y}_j||$. When $y_i$ is close to $y_j$ in the low-dimensional space (i.e., $||\vec{y}_i-\vec{y}_j||$ is small), there is a large probability of observing a binary edge between the two vertices. In reality, many probabilistic functions can be used such as $f(x)=\frac{1}{1+ax^2}$ or $f(x) = \frac{1}{1+\exp(x^2)}$. We compare different probabilistic functions in Section~\ref{sec::experiments}. 

Eqn.~\eqref{eqn::edge_probability} only defines the probability of observing a binary edge between a pair of vertices. To further extend it to general weighted edges, we define the likelihood of observing a weighted edge $e_{ij}=w_{ij}$ as follows:
\begin{equation}
\label{eqn::weighted_edge_probability}
P(e_{ij}=w_{ij}) =P(e_{ij}=1)^{w_{ij}}.
\end{equation}

With the above definition, given a weighted graph $G=(V,E)$, the likelihood of the graph can be calculated as:
\begin{equation}
	\label{eqn::obj}
	\begin{aligned}
	O=&\prod_{(i,j)\in E} p(e_{ij}=1)^{w_{ij}} \prod_{(i,j)\in \bar{E}} (1-p(e_{ij}=1))^{\gamma}\\
	\propto&\sum_{(i,j)\in E} w_{ij} \log p(e_{ij}=1)+ \sum_{(i,j)\in \bar{E}} \gamma\log (1-p(e_{ij}=1)),
	\end{aligned}
\end{equation}
in which $\bar{E}$ is the set of vertex pairs that are not observed and $\gamma$ is an unified weight assigned to the negative edges. The first part of Eqn.~\eqref{eqn::obj} models the likelihood of the observed edges, and by maximizing this part similar data points will keep close together in the low-dimensional space; the second part models the likelihood of all the vertex pairs without edges, i.e., negative edges. By maximizing this part, dissimilar data will be far away from each other. By maximizing the objective~\eqref{eqn::obj}, both goals can be achieved.

\noindent\textbf{Optimization.} Directly maximizing Eqn.~\eqref{eqn::obj} is computationally expensive, as the number of negative edges is quadratic to the number of nodes. Inspired by the negative sampling techniques~\cite{mikolov2013distributed}, instead of using all the negative edges, we randomly sample some negative edges for model optimization. For each vertex $i$, we randomly sample some vertices $j$ according to a noisy distribution $P_n(j)$ and treat $(i,j)$ as the negative edges. We used the noisy distribution in~\cite{mikolov2013distributed}: $P_n(j)\propto d_j^{0.75}$, in which $d_j$ is the degree of vertex $j$. 
Letting $M$ be the number of negative samples for each positive edge, the objective function can be redefined as: 
\begin{eqnarray}
	\label{eqn::obj_ns}
\nonumber	O & = & \sum_{(i,j)\in E} w_{ij}\big(\log p(e_{ij}=1) + \\
	& & \sum_{k=1}^M E_{j_k\sim P_n(j)} \gamma \log (1-p(e_{ij_k}=1))\big).
\end{eqnarray}

A straightforward approach to optimize Eqn.~\eqref{eqn::obj_ns} is stochastic gradient descent, which is problematic however. This is because when sampling an edge $(i,j)$ for model updating, the weight of the edge $w_{ij}$ will be multiplied into the gradient. When the values of the weights diverge (e.g., ranging from 1 to thousands), the norms of the gradient also diverge, in which case it is very difficult to choose an appropriate learning rate. We adopt the approach of edge sampling proposed in our previous paper~\cite{tang2015line}. We randomly sample the edges with the probability proportional to their weights and then treat the sampled edges as binary edges. With this edge sampling technique, the objective function remains the same and the learning process will not be affected by the variance of the weights of the edges.

To further accelerate the training process, we adopt the asynchronous stochastic gradient descent, which is very efficient and effective on sparse graphs~\cite{recht2011hogwild}. The reason is that when different threads sample different edges for model updating, as the graph is very sparse, the vertices of the sampled edges in different threads seldom overlap, i.e., the embeddings of the vertices or the model parameters usually do not conflict across different threads. 

For the time complexity of the optimization, each stochastic gradient step takes $O(sM)$, where $M$ is the number of negative samples and $s$ is the dimension of low-dimensional space (e.g., 2 or 3). In practice, the number of stochastic gradient steps is typically proportional to the number of vertices $N$. Therefore, the overall time complexity is $O(sMN)$, which is linear to the number of nodes $N$. 

\section{Experiments}
\label{sec::experiments}
We evaluate the efficiency and effectiveness of the LargeVis both quantitatively and qualitatively. In particular, we separately evaluate the performance of the proposed algorithms for constructing the KNN graph and visualizing the graph. 

\begin{table}
	\centering
	\caption{Statistics of the data sets.}
	\label{tab::data_statistics}
	\scalebox{0.8}{	
	\begin{tabular}{c|c|c|c}\hline
		Data Set& \# data & \# dimension & \# categories \\ \hline
		20NG&18,846 &100 &20  \\ \hline
		MNIST&70,000 &784 &10  \\ \hline
		WikiWord&836,756 &100 & -  \\ \hline
		WikiDoc&2,837,395 & 100& 1,000 \\ \hline
		CSAuthor&1,854,295 &100 & - \\ \hline
		DBLPPaper&  1,345,560     & 100 & - \\ \hline
		LiveJournal&3,997,963 &100 &5,000  \\ \hline										
	\end{tabular}
	}
\end{table}

\subsection{Data Sets}
We select multiple large-scale and high-dimensional data sets of various types including text (words and documents), images, and networks including the following: 
\begin{itemize}
	\item 20NG: the widely used text mining data set 20newsgroups\footnote{Available at \url{http://qwone.com/~jason/20Newsgroups/}}. We treat each article as a data point.
	\item MNIST: the handwritten digits data set\footnote{Available at \url{http://yann.lecun.com/exdb/mnist/}}. Each image is treated as a data point.
	\item WikiWord: the vocabulary in the Wikipedia articles\footnote{\url{https://en.wikipedia.org/wiki/Wikipedia:Database_download}} (words with frequency less than 15 are removed). Each word is a data point.
	\item WikiDoc: the entire set of English Wikipedia articles (articles containing less than 1000 words are removed). Each article is a data point. We label the articles with the top 1,000 Wikipedia categories and label all the other articles with a special category named ``others.''
	\item CSAuthor: the co-authorship network in the computer science domain, collected from Microsoft Academic Search. Each author is a data point.
	\item DBLPPaper: the heterogeneous networks of authors, papers, and conferences in the DBLP data\footnote{Available at \url{http://dblp.uni-trier.de/xml/}}. Each paper is a data point.
	\item LiveJournal: the LiveJournal social network\footnote{Available at \url{https://snap.stanford.edu/data/}}. Every node is labeled with the communities it belongs to, if it is one of the most popular 5,000 communities, or with a special category named ``others.'' 
\end{itemize}

Note that although the original data sets all come with variety numbers of dimensions (e.g., size of the vocabulary for text documents), for comparison purposes we represent them with a fixed number of dimensions (e.g., 100) before applying any visualization techniques. This step is not required for LargeVis in practice, but learning an intermediate representation of the data can improve (e.g., smooth) the similarity structure of the original data. There are quite a few efficient embedding learning techniques (such as Skipgram~\cite{mikolov2013distributed} and LINE~\cite{tang2015line}), the computational cost of which will not be a burden of the visualization. Specifically, the representations of nodes in network data are learned through the LINE; the representations of words are learned through the LINE using a simple co-occurrence network; and the representations of documents are simply taken as the averaged vectors of the words in the documents. 
The vector representation of the image data is already provided from the source, so we do not further learn a new embedding. 

\begin{figure*}[htbp]
	\centering
	\subfigure[WikiWord]{
		\label{fig::efficiency_vs_accuracy_knng_wiki_word}
		\includegraphics[width=0.22\textwidth]{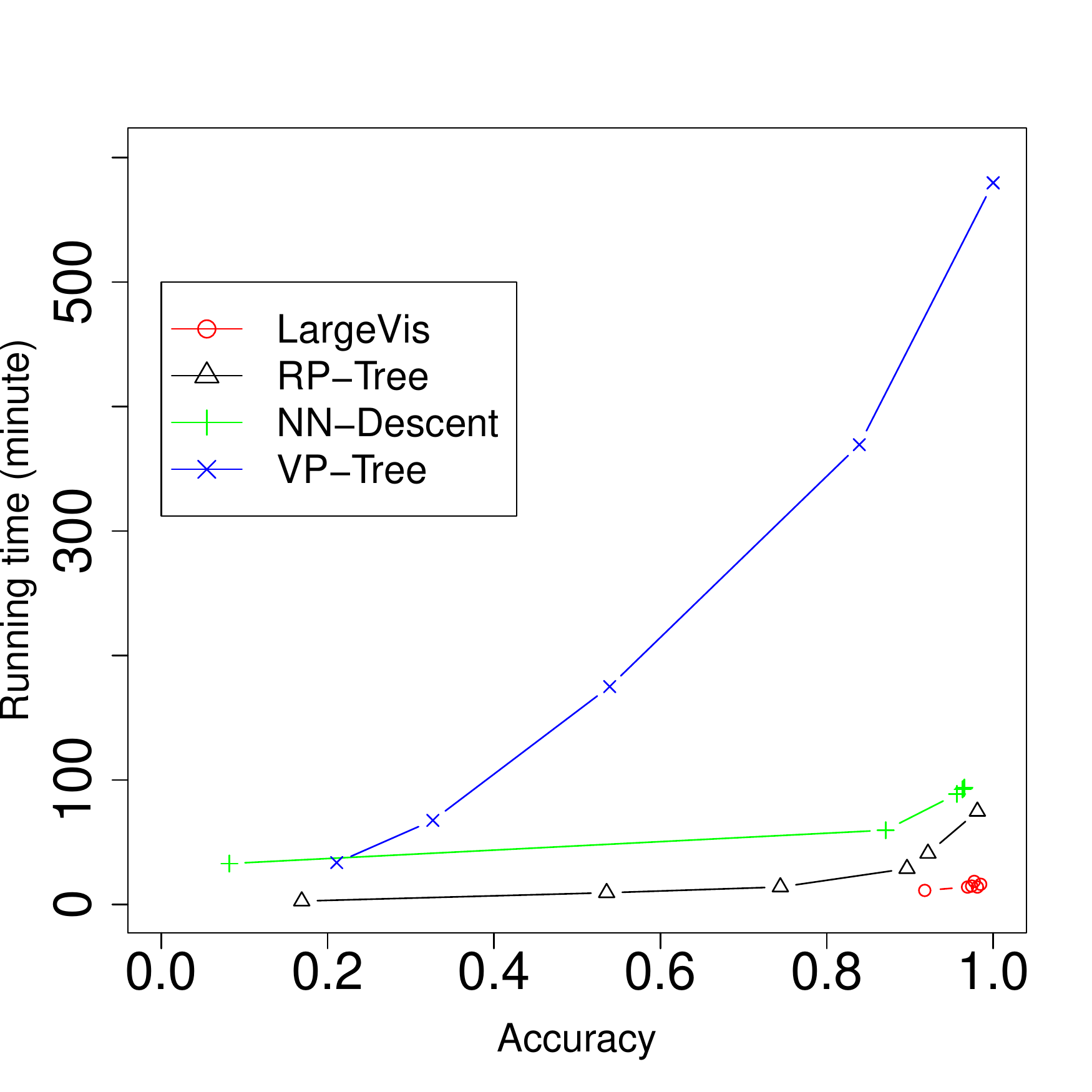}
	}
	\subfigure[WikiDoc]{
		\label{fig::efficiency_vs_accuracy_knng_wiki_doc}
		\includegraphics[width=0.22\textwidth]{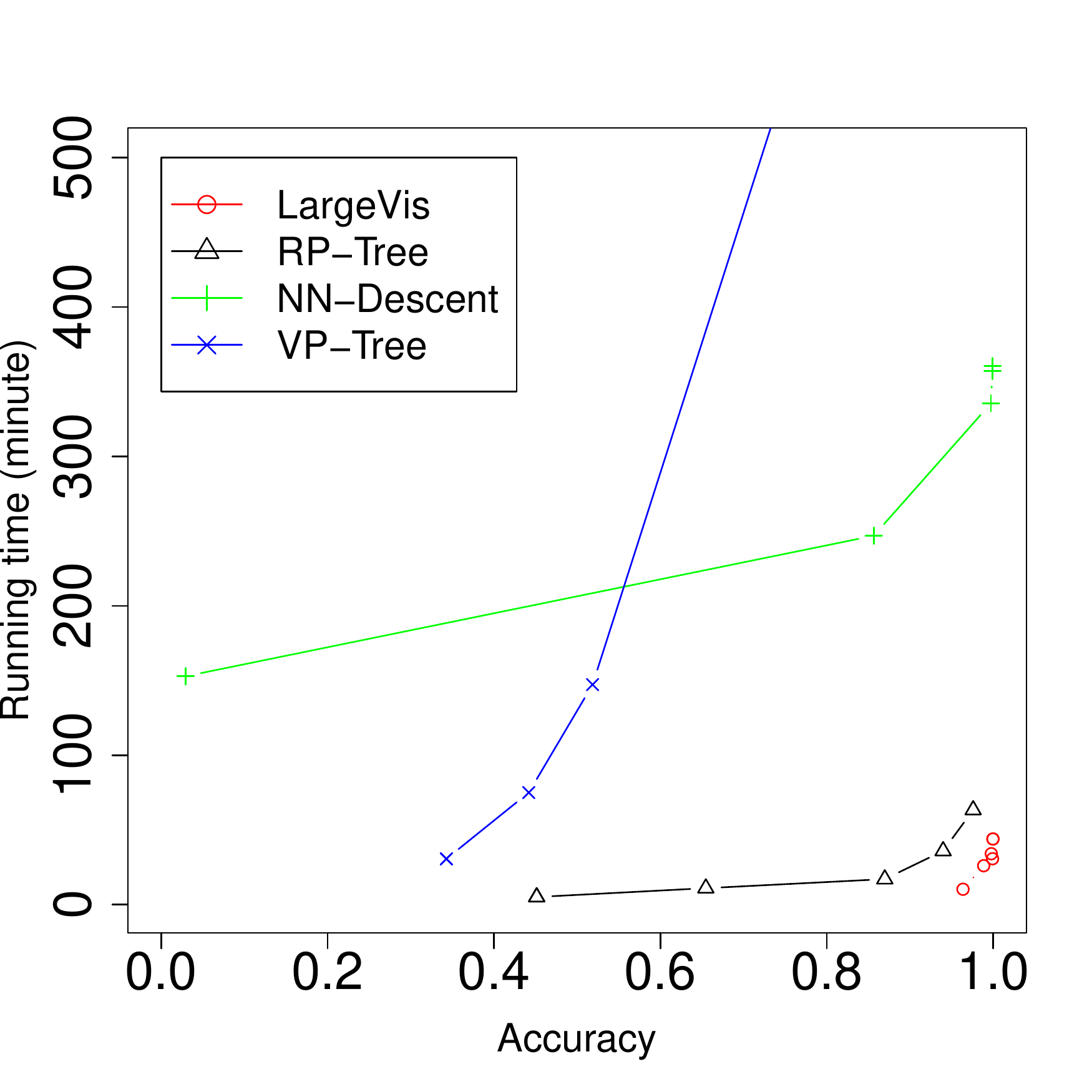}
	}
	\subfigure[LiveJournal]{
		\label{fig::efficiency_vs_accuracy_knng_lj}
		\includegraphics[width=0.22\textwidth]{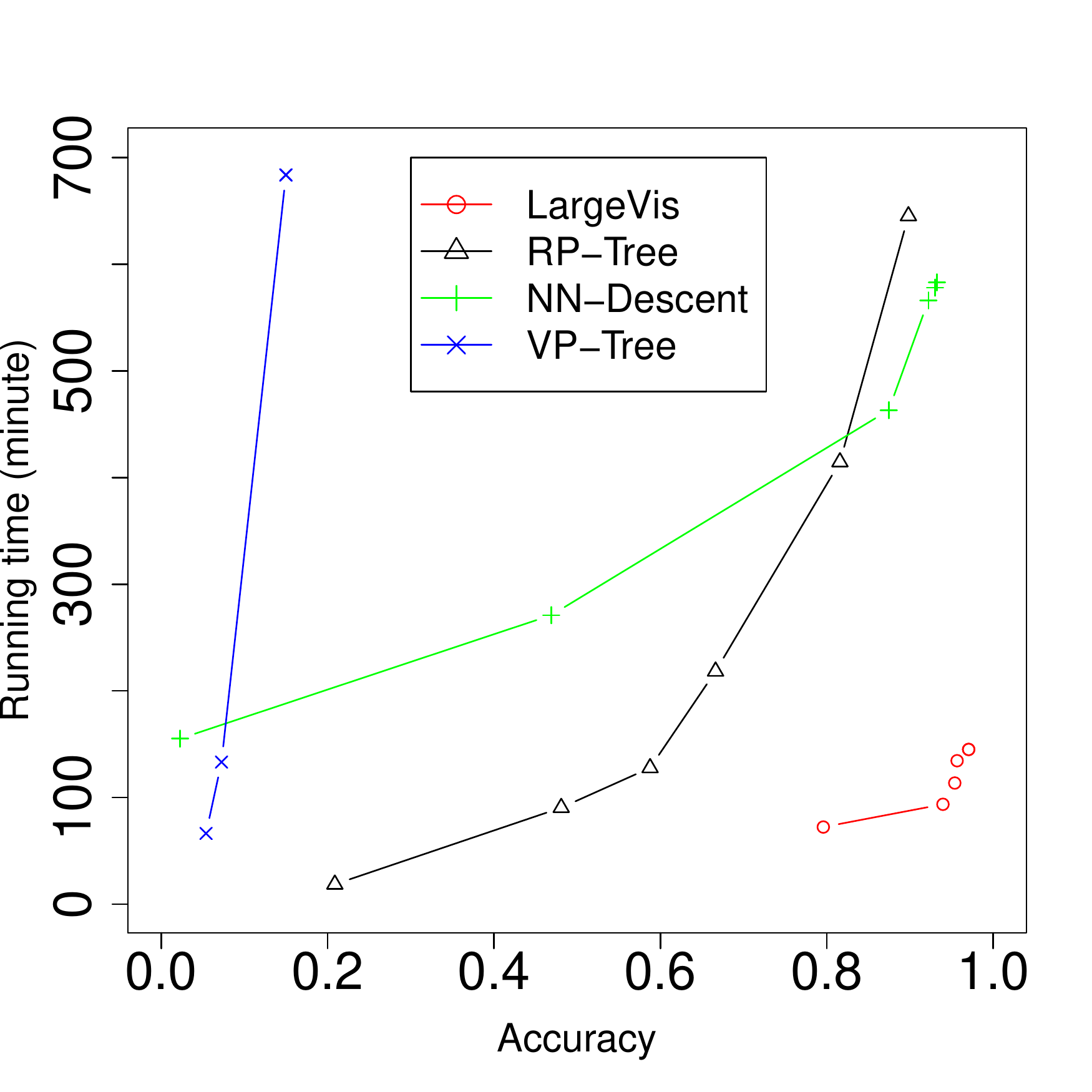}
	}
	\subfigure[CSAuthor]{
		\label{fig::efficiency_vs_accuracy_knng_lj}
		\includegraphics[width=0.22\textwidth]{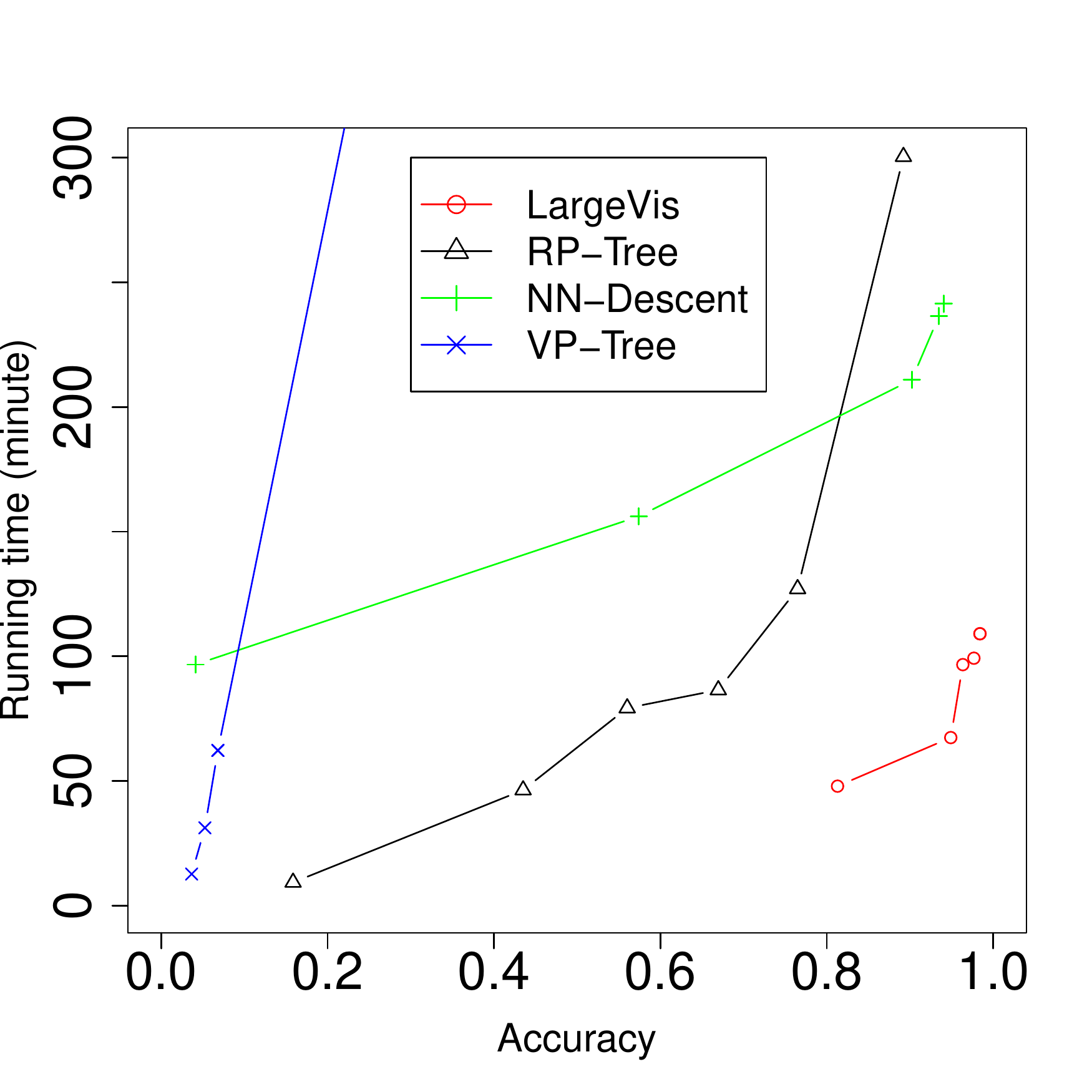}
	}			
	\caption{Running time v.s. Accuracy of KNN graph construction. The lower right corner indicates optimal performance. LargeVis outperforms the vantage-point tree and other state-of-the-art methods.}
	\label{fig::results_knng}
\end{figure*}

We summarize the statistics of the above data sets in Table~\ref{tab::data_statistics}. Next, we report the results of KNN graph construction and graph visualization respectively. All the following results are executed on a machine with 512GB memory, 32 cores at 2.13GHz. When multiple threads are used, the number of threads is always 32. For visualization purposes, in all the experiments, we learn a 2D layout of the data.

\subsection{Results on KNN Graph Construction}

We first compare the performance of different algorithms for K-nearest neighbor graph construction, including:
\begin{itemize}
	\item Random Projection Trees~\cite{dasgupta2008random}.  We use the implementation of random projection trees in the Annoy\footnote{https://github.com/spotify/annoy} system. 
	\item Vantage-point trees~\cite{yianilos1993data}. This is the approach used by the t-SNE. 
	\item NN-Descent~\cite{dong2011efficient}. This is a representative neighbor exploring technique.
	\item LargeVis. Our proposed technique by improving random projection trees with neighbor exploration.
\end{itemize}

Fig.~\ref{fig::results_knng} compares the performance of different algorithms for KNN graph construction. The number of neighbors for each data point is set as 150. For each algorithm, we try different values of its parameters, resulting in a curve of running time over accuracy (i.e., the percentage of data points that are truly K-nearest neighbors of a node). Some results of the vantage-point trees could not be shown as the values are too large. For LargeVis, only one iteration of neighbor exploring is conducted. Overall, the proposed graph construction algorithm consistently achieves the best performance (the shortest running time at the highest accuracy) on all the four data sets, and the vantage-point trees perform the worst. On the WikiDoc data set, which contains around 3 million data points, our algorithm takes only 25 minutes to achieve 95\% accuracy while vantage-point trees take 16 hours, which is 37 times slower. Compared to the original random projection trees, the efficiency gain is also salient. On some data sets, e.g., LiveJournal and CSAuthor, it is very costly to construct a KNN graph at a 90\% accuracy through random projection trees. However, with the neighbor exploring techniques, the accuracy of the graph significantly improves to near perfection. 

\begin{figure}[htbp]
	\centering
	\subfigure[WikiDoc]{
		\label{fig::efficiency_vs_accuracy_knng_lj}
		\includegraphics[width=0.22\textwidth]{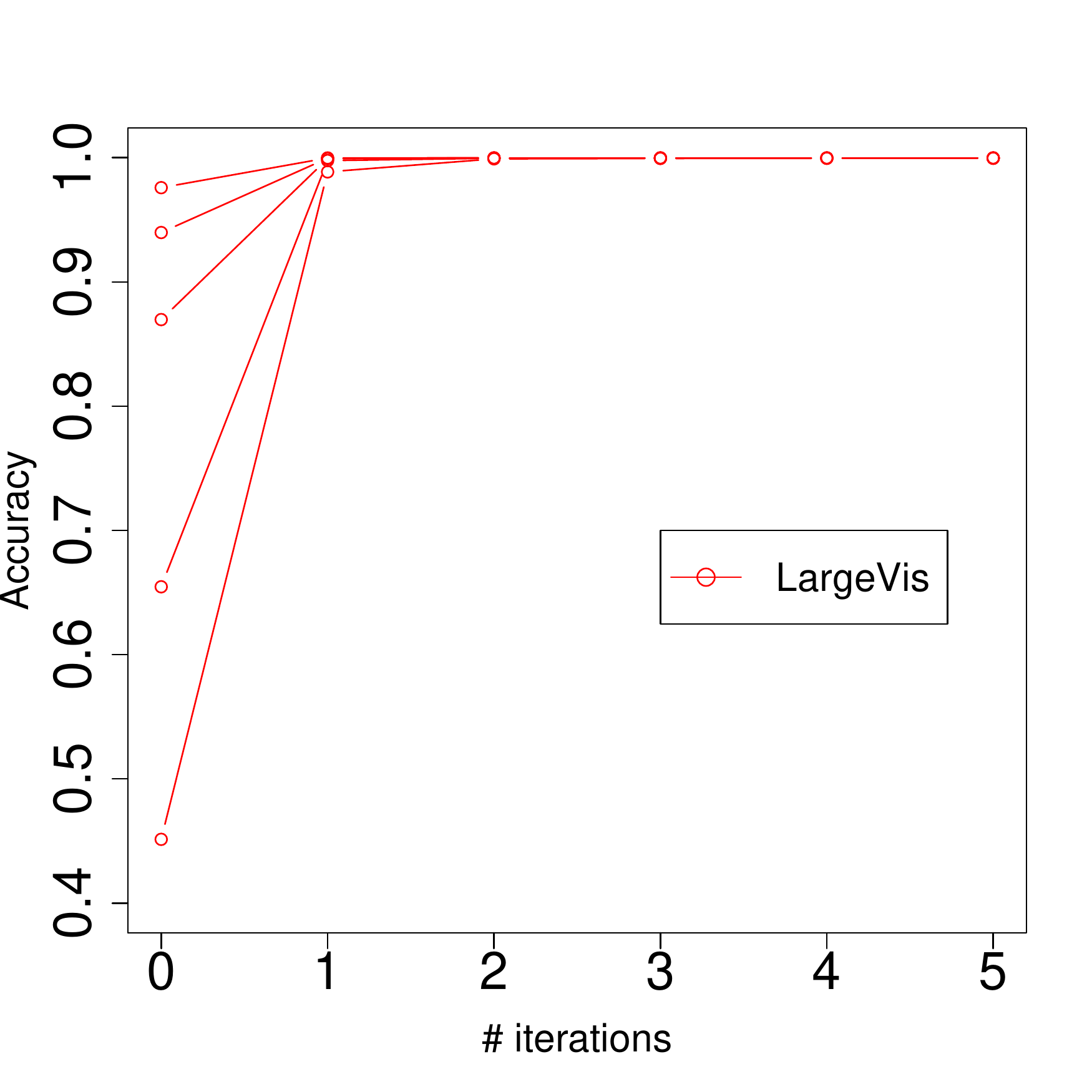}
	}
	\subfigure[LiveJournal]{
		\label{fig::efficiency_vs_accuracy_knng_lj}
		\includegraphics[width=0.22\textwidth]{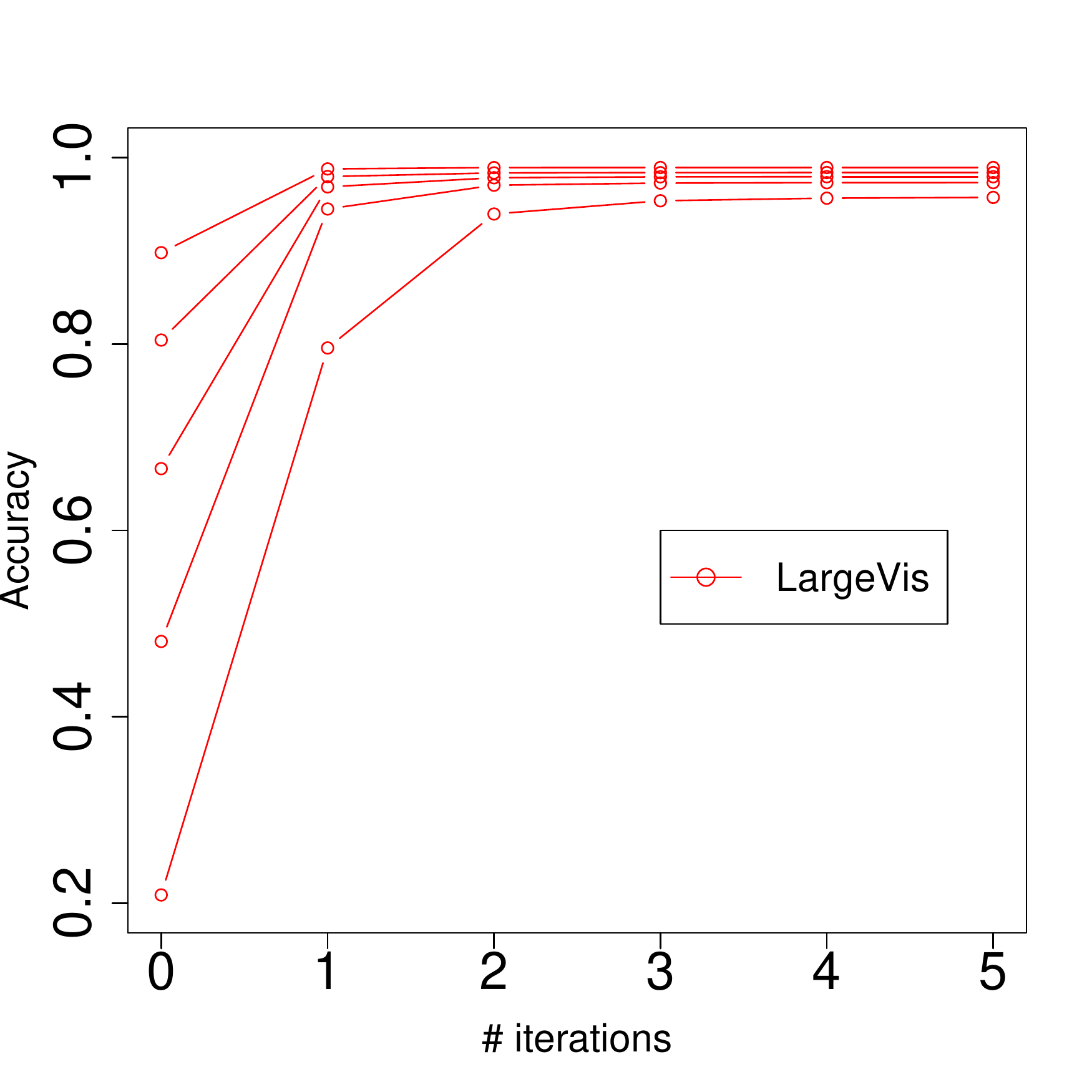}
	}
	\caption{Accuracy of KNN Graph w.r.t number of iterations of neighbor exploring in LargeVis. Curves correspond to initializing KNN Graphs at different levels of accuracy. }
	\label{fig::results_knng_propagations}
\end{figure}

How many iterations of neighbor exploring are required for LargeVis to achieve a good accuracy? Fig.~\ref{fig::results_knng_propagations} presents the results of the accuracy of KNN Graph w.r.t the number of iterations of neighbor exploring. We initialize KNN graphs with different levels of accuracy, constructed with different numbers of random projection trees. Neighbor exploring is very effective. On WikiDoc, the accuracy of the approximate KNN graph improves from 0.4 to almost 1 with only one iteration of neighbor exploring. On LiveJounal, at most three iterations are needed to achieve a very high accuracy, even if starting from a very inaccurate KNN graph. Similar results are also observed in other data sets. 

Our proposed algorithm for KNN graph construction is very efficient, easily scaling to millions of data points with hundreds of dimensions. This solves the computational bottleneck of many data visualization techniques. Next, we compare algorithms that visualize the KNN graphs. All visualization algorithms use the same KNN graphs constructed by LargeVis as input, setting the perplexity to 50 and the number of neighbors for each data point to 150.

\subsection{Graph Visualization}
We compare the following graph visualization algorithms:
\begin{itemize}
	\item Symmetric SNE~\cite{hinton2002stochastic}. The approach of symmetric stochastic neighbor embedding. To scale it up for large graphs, the Barnes-Hut algorithm~\cite{van2014accelerating} is used for acceleration.
	\item t-SNE~\cite{van2014accelerating}. The state-of-the-art approach for visualizing high-dimensional data, also accelerated through the Barnes-Hut algorithm. 
	\item LINE~\cite{tang2015line}. A large-scale network/graph embedding method. Although not designed for visualization purposes, we directly learn a 2-dimensional embedding. First-order proximity~\cite{tang2015line} is used. 
	\item LargeVis. Our proposed technique for graph visualization introduced in Section~\ref{sec::visualization}. 
\end{itemize}

\noindent \textbf{Model Parameters and Settings.} For the model parameters in SNE and t-SNE, we set $\theta=0.5$ and the number of iterations to $1,000$, which are suggested by~\cite{van2014accelerating}. For the learning rate of t-SNE, we find the performance is very sensitive w.r.t. different values and the optimal values on different data sets vary significantly. We report the results with the default learning rate 200 and the optimal values respectively. For both LINE and LargeVis, the size of mini-batches is set as 1; the learning rate is set as $\rho_t=\rho(1-t/T)$, where $T$ is the total number of edge samples or mini-batches. Different values of initial learning rate is used by LINE and LargeVis: $\rho_0=0.025$ in LINE and $\rho_0=1$ in LargeVis. The number of negative samples is set as 5 and $\gamma$ is set as 7 by default. The number of samples or mini-batches $T$ can be set proportional to the number of nodes. In practice, a reasonable number of $T$ for 1 million nodes is 10K million. The LINE and LargeVis can be naturally parallelized through asynchronously stochastic gradient descent. We also parallelize Symmetric SNE and t-SNE by assigning different nodes into different threads in each full batch gradient descent.   

\noindent \textbf{Evaluation.} The evaluation of data visualization is usually subjective. Here we borrow the approach adopted by the t-SNE to evaluate the visualizations quantitatively~\cite{van2014accelerating}. We use a KNN classifier to classify the data points based on their low-dimensional representations. The intuition of this evaluation methodology is that a good visualization should be able to preserve the structure of the original data as much as possible and hence yield a high classification accuracy with the low-dimensional representations.  

\subsubsection{Comparing Different Probabilistic Functions}

\begin{figure}[htdb!]
	\centering
	\subfigure[WikiDoc]{
		\label{fig::efficiency_vs_accuracy_knng_xx}
		\includegraphics[width=0.22\textwidth]{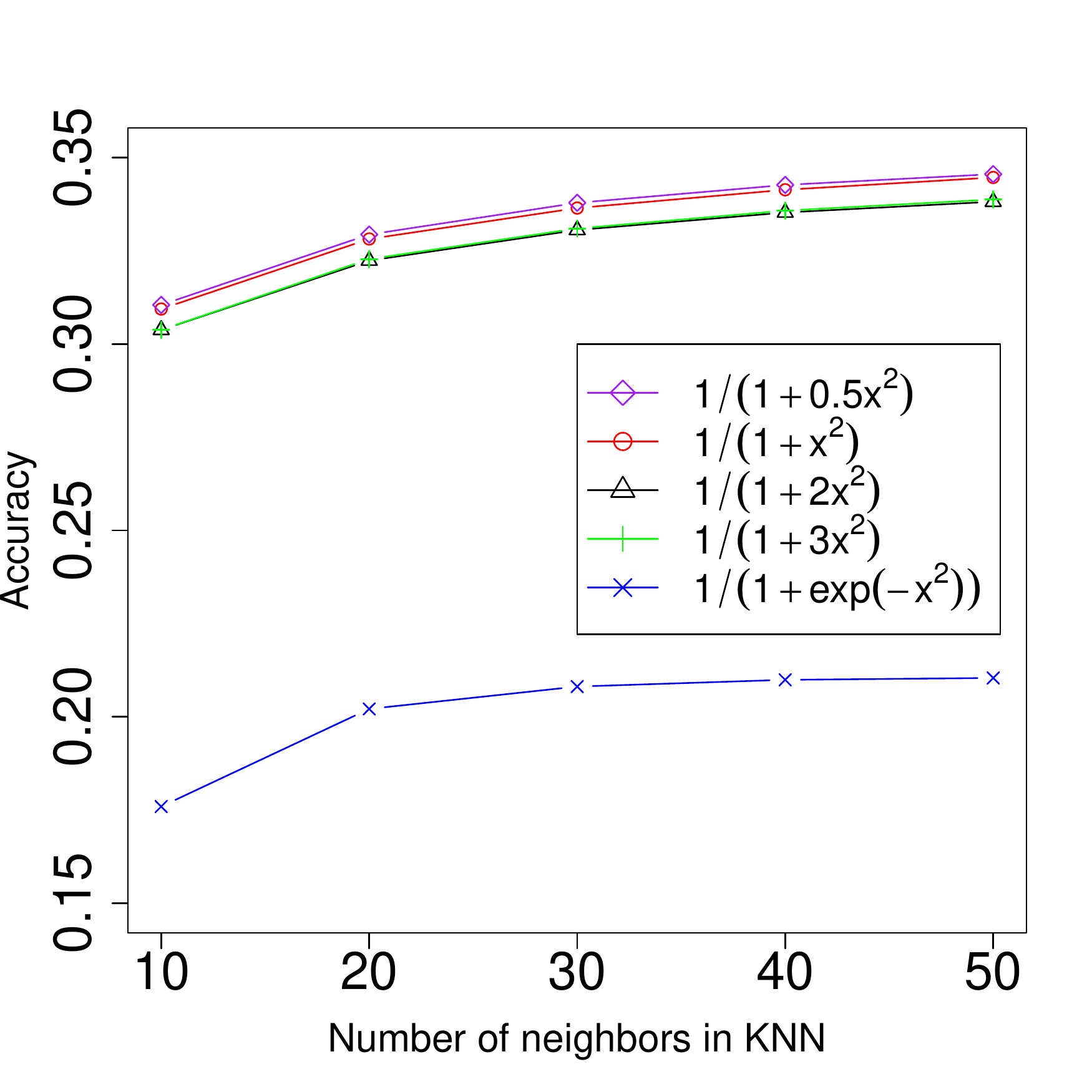}
	}	
	\subfigure[LiveJournal]{
		\label{fig::efficiency_vs_accuracy_knng_xx}
		\includegraphics[width=0.22\textwidth]{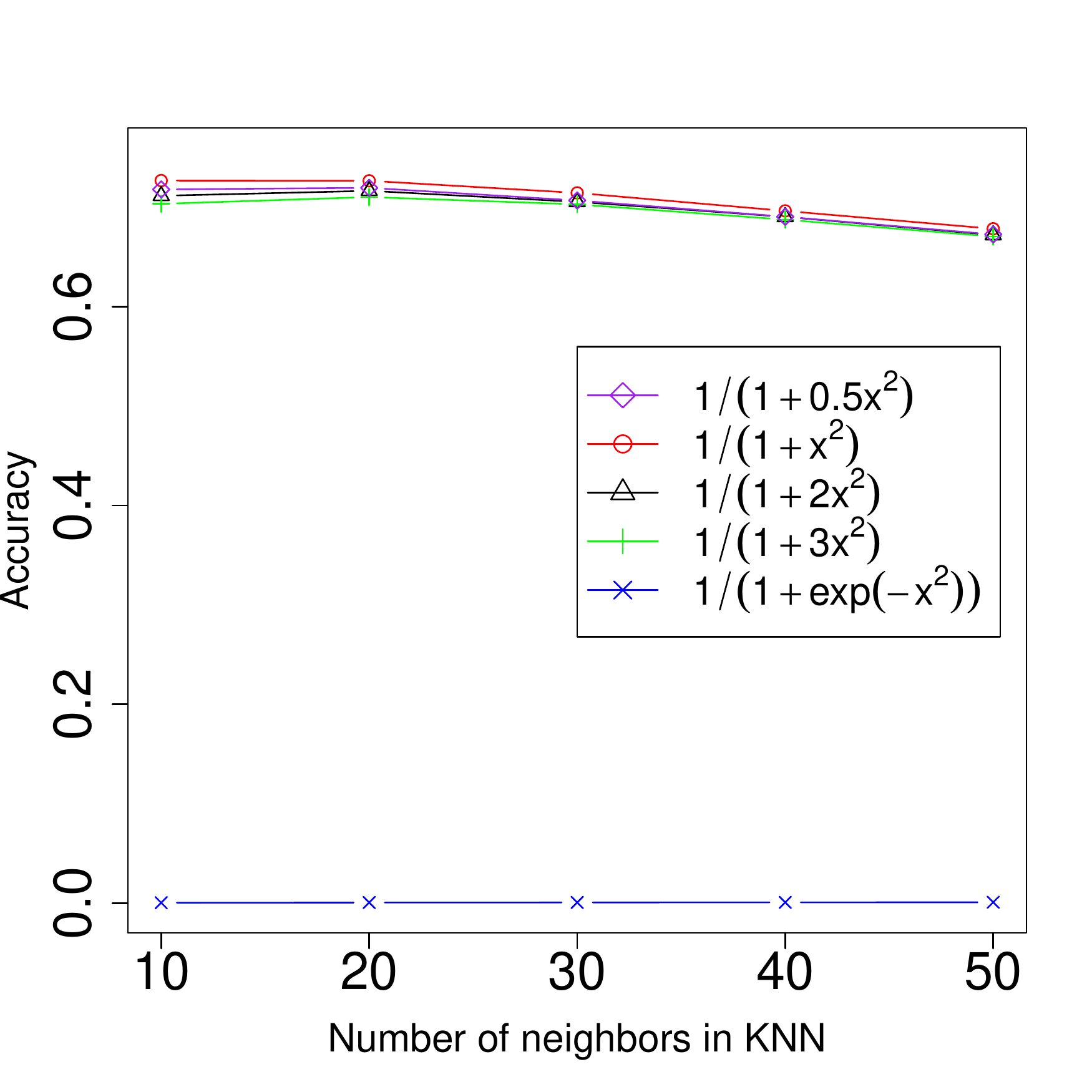}
	}					
	\caption{Comparing different probabilistic functions.}
	\label{fig::results_probabilistic_functions}
\end{figure}

We first compare different probabilistic functions in Eq.~\eqref{eqn::edge_probability}, which define the probability of observing a binary edge between a pair of vertices based on the distance of their low-dimensional representations. We compare functions $f(x) = \frac{1}{1+ax^2}$ and $f(x) = \frac{1}{1+\exp(-x^2)}$ with various values of $a$. Fig.~\ref{fig::results_probabilistic_functions} presents the results on the WikiDoc and the LiveJournal data sets. We can see among all probabilistic functions, $ f(x) = \frac{1}{1+x^2}$ achieves the best result. This probability function specifies a long-tailed distribution, therefore can also solve the ``crowding problem'' according to~\cite{van2008visualizing}. In the following experiments, we always use $f(x) = \frac{1}{1+x^2}$. 

\subsubsection{Results on Different Data Sets}

\begin{figure*}[htdb!]
	\centering
	\subfigure[20NG]{
		\label{fig::efficiency_vs_accuracy_knng_wiki_word}
		\includegraphics[width=0.22\textwidth]{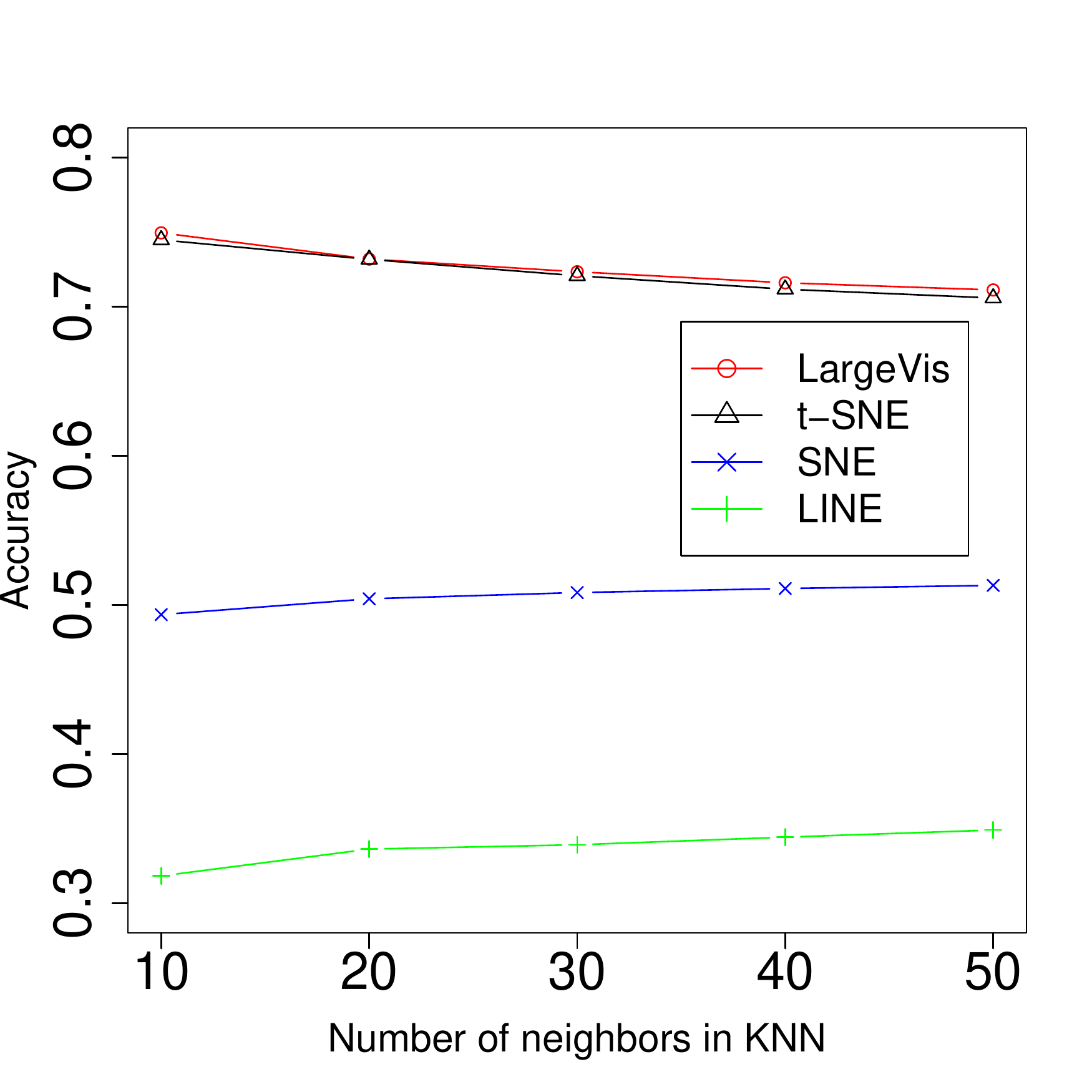}
	}
	\subfigure[MNIST]{
		\label{fig::efficiency_vs_accuracy_knng_wiki_doc}
		\includegraphics[width=0.22\textwidth]{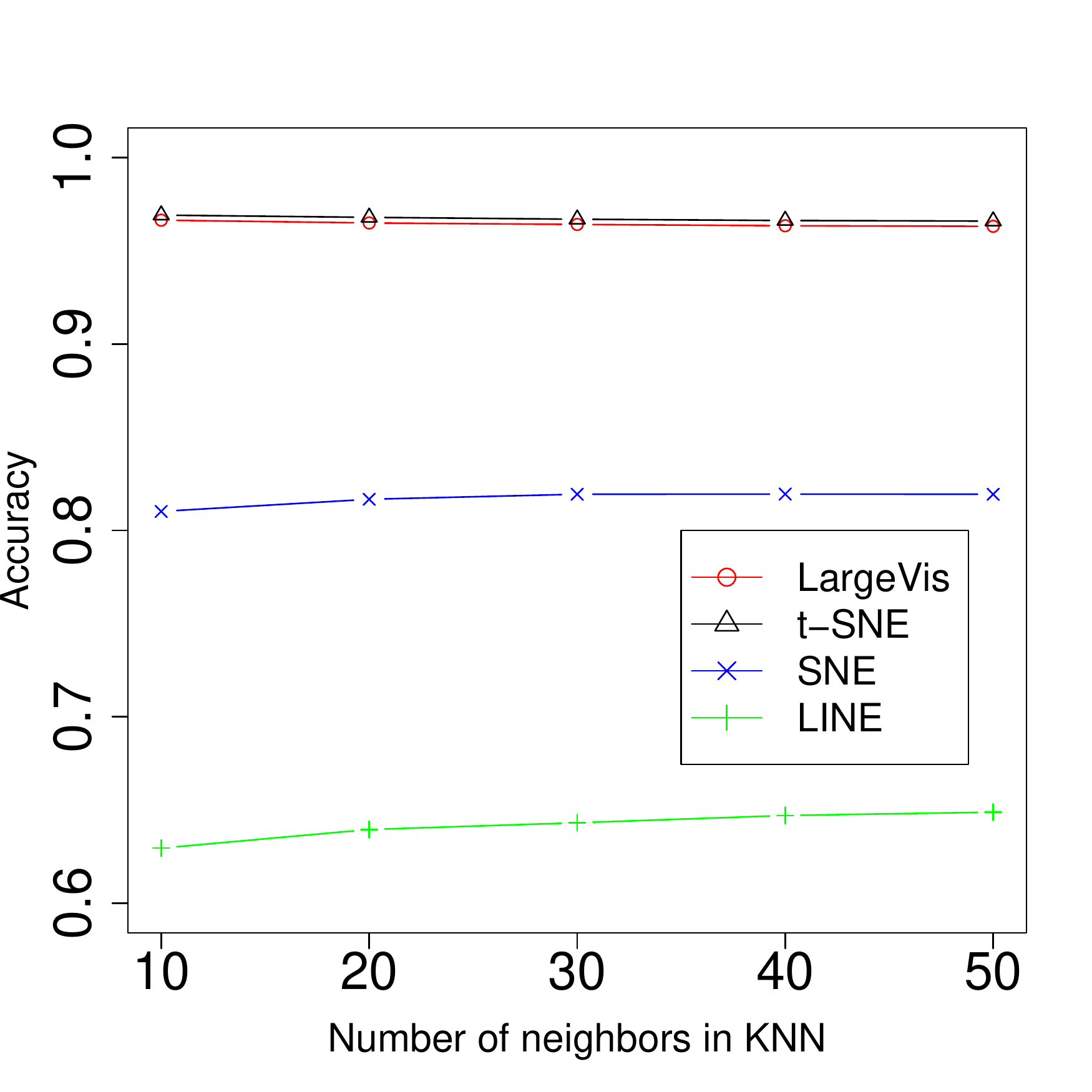}
	}
	\subfigure[WikiDoc]{
		\label{fig::efficiency_vs_accuracy_knng_xx}
		\includegraphics[width=0.22\textwidth]{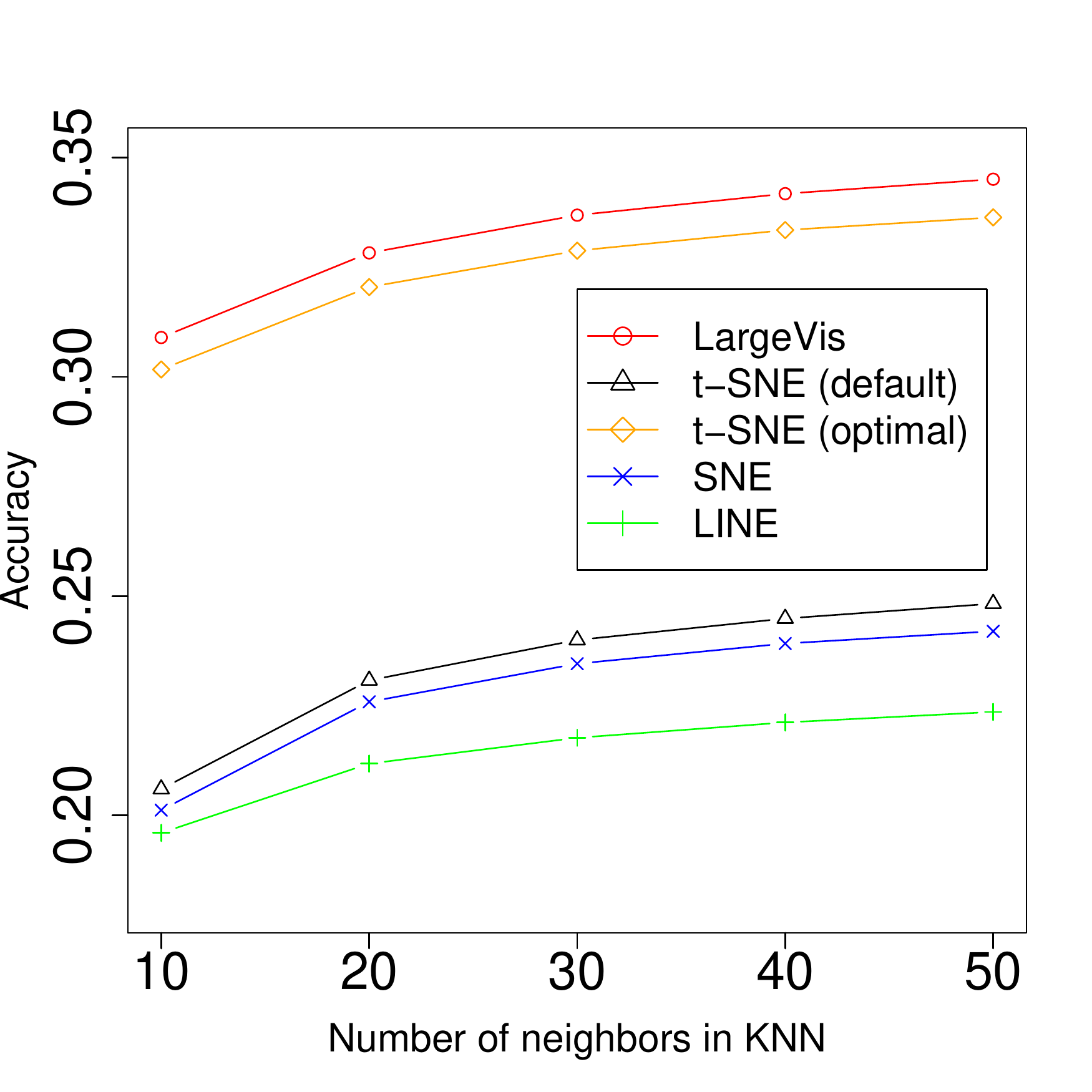}
	}	
	\subfigure[LiveJournal]{
		\label{fig::efficiency_vs_accuracy_knng_xx}
		\includegraphics[width=0.22\textwidth]{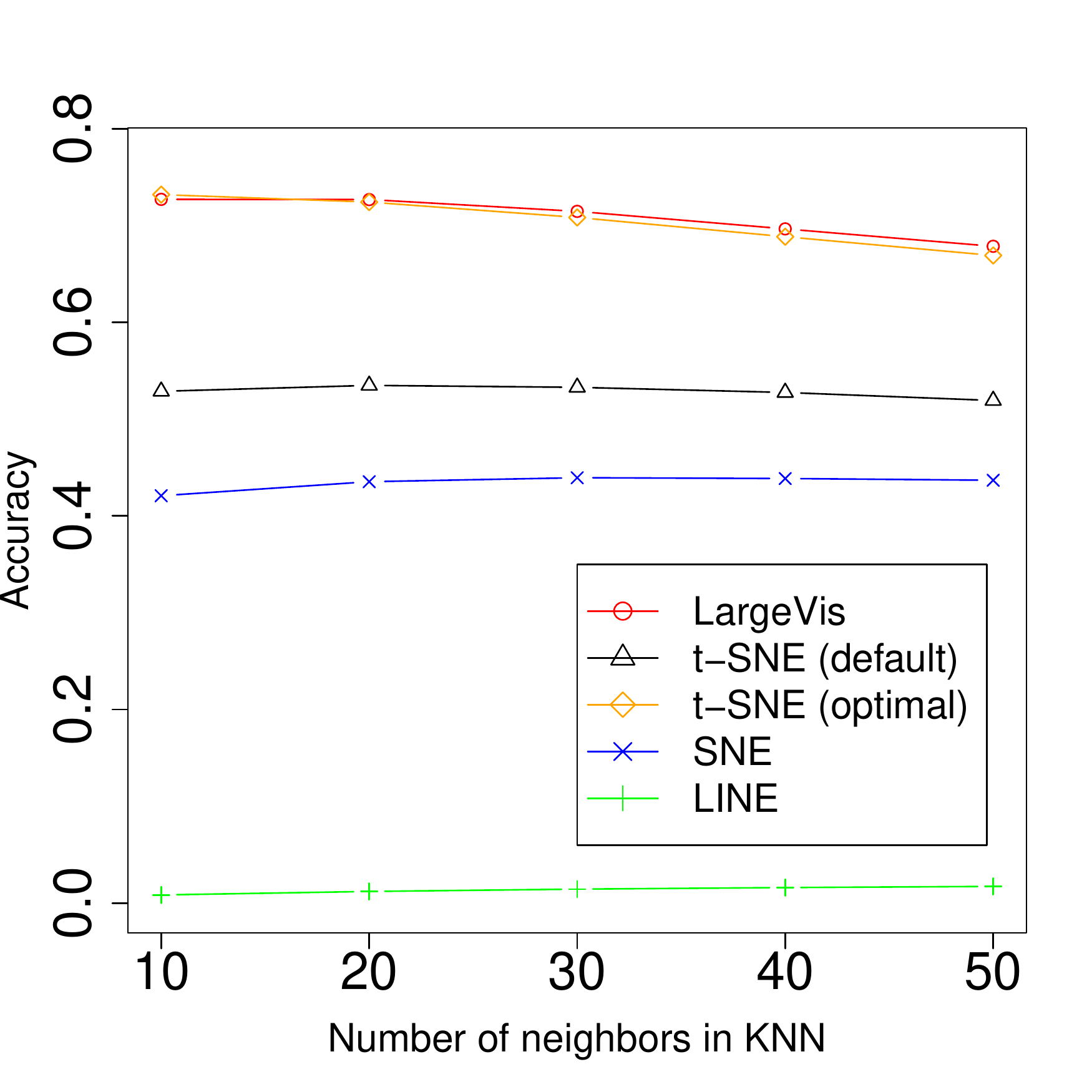}
	}					
	\caption{Performance of classifying data points according to 2D representations using the K-nearest neighbor classifier. Overall, the LargeVis is more effective or comparable to t-SNE with optimal learning rates, significantly outperforming t-SNE with the recommended learning rate (1,000) on large data sets. The optimal learning rates of t-SNE vary significantly on different data sets, ranging from around 200 (on 20NG and MNIST) to 2,500 (on WikiDoc) and 3,000 (on LiveJournal), which are very expensive to search on large data sets. Even with the optimal parameters, t-SNE is inferior to LargeVis which simply uses default parameters on all data sets. }
	\label{fig::results_visualization}
\end{figure*}

We compare the efficiency and effectiveness of different visualization algorithms. Fig.~\ref{fig::results_visualization} compares the classification accuracy with the K-nearest neighbor classifier by using the low-dimensional representations as features. For the KNN classifier, different numbers of neighbors are tried. For t-SNE, both the results with default learning rate 200 and the optimal learning rate tuned thorough exhaustively search are reported. On the small data sets 20NG and MNIST, which contain less than 100,000 data points, the default learning rate of t-SNE yields optimal performance, which is comparable to LargeVis. However, on the large data sets WikiDoc and LiveJournal, which contain millions of data points, the LargeVis is more effective or comparable to the t-SNE with optimal learning rates, significantly outperforming t-SNE with default learning rate. However, empirically tuning the learning rate of t-SNE requires repeatedly training, which is very time consuming on the large data sets. The optimal learning rates of t-SNE on different data sets vary significantly. On the small data sets 20NG and MNIST, the optimal learning rate is around 200, while on the large data sets WikiDoc and LiveJournal, the optimal values become as large as 3000. Comparing to t-SNE, the performance of LargeVis is very stable w.r.t the learning rate, the default value of which can be applied to various data sets with different sizes. We also notice that the performance of the LINE is very bad, showing that an embedding learning method is not appropriate for data visualization as is. 

\begin{table*}
	\centering
	\caption{Comparison of running time (hours) in graph visualization between the t-SNE and LargeVis.}
	\label{tab::Running_time}
	\begin{tabular}{c|c|c|c|c|c|c|c|c}  \hline
		Algorithm  & 20NG& MNIST & WikiWord & WikiDoc & LiveJournal & CSAuthor & DBLPPaper \\ \hline
		t-SNE &0.12		&0.41 & 9.82	&45.01	&70.35	&28.33	&18.73 \\ \hline
		LargeVis &0.14	&0.23	&2.01		&5.60		&9.26		&4.24		&3.19 \\ \hline
		Speedup Rate &0 	&0.7	&3.9		&7		&6.6		&5.7		&4.9 \\ \hline
	\end{tabular}
\end{table*}

Table~\ref{tab::Running_time} compares the running time of t-SNE and LargeVis for graph visualization. On the small data sets 20NG and MNIST, the two algorithms perform comparable to each other. However, on the large data sets, the LargeVis is much more efficient than the t-SNE. Specially, on the largest data set LiveJournal, which contains 4 million data points, the LargeVis is 6.6 times faster than the t-SNE.  

\subsubsection{Performance w.r.t. Data Size}

\begin{figure*}[htdb!]
	\centering
	\subfigure[Accuracy (WikiDoc)]{
		\label{fig::accuracy_vs_size_ww}
		\includegraphics[width=0.22\textwidth]{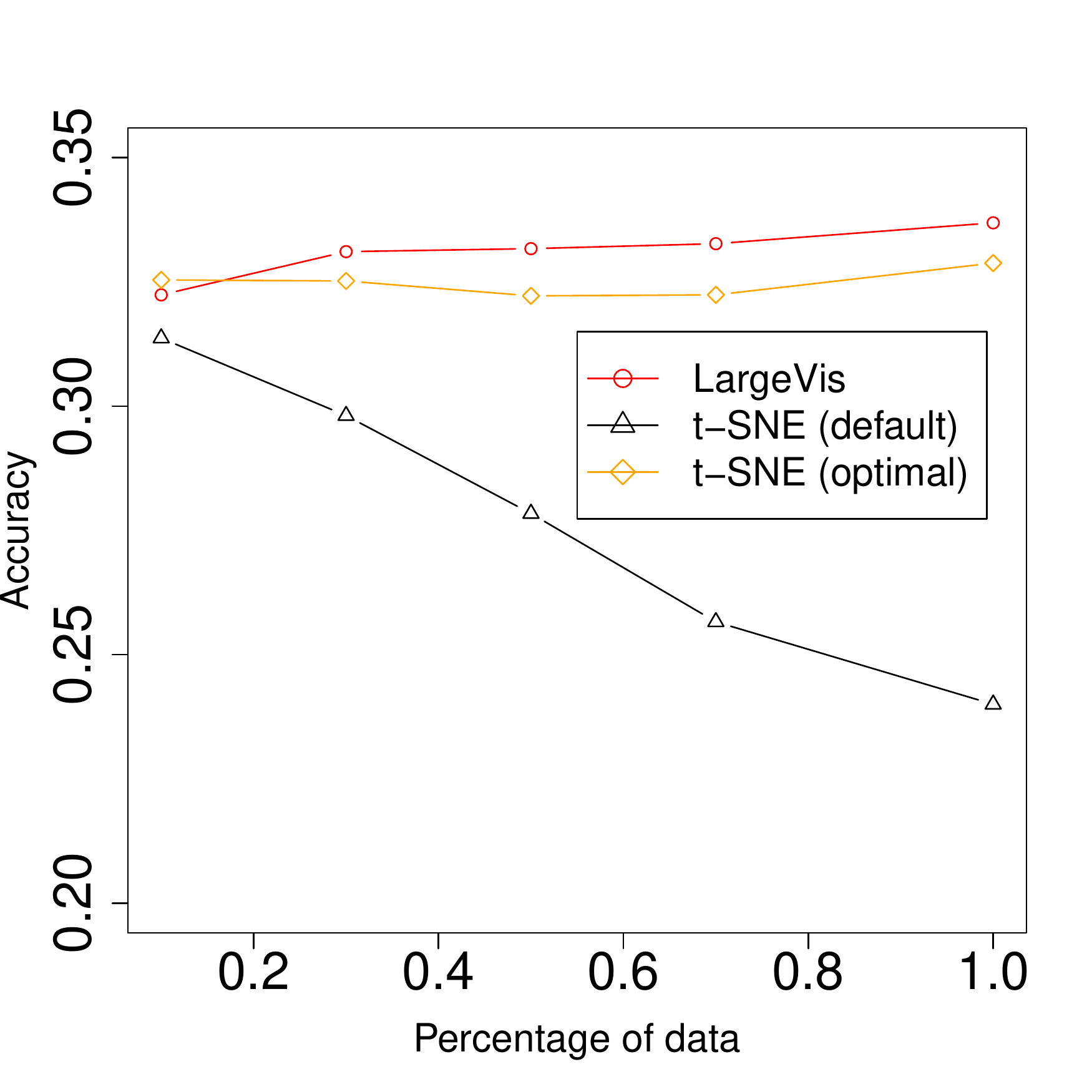}
	}
	\subfigure[Accuracy (LiveJournal)]{
		\label{fig::accuracy_vs_size_lj}
		\includegraphics[width=0.22\textwidth]{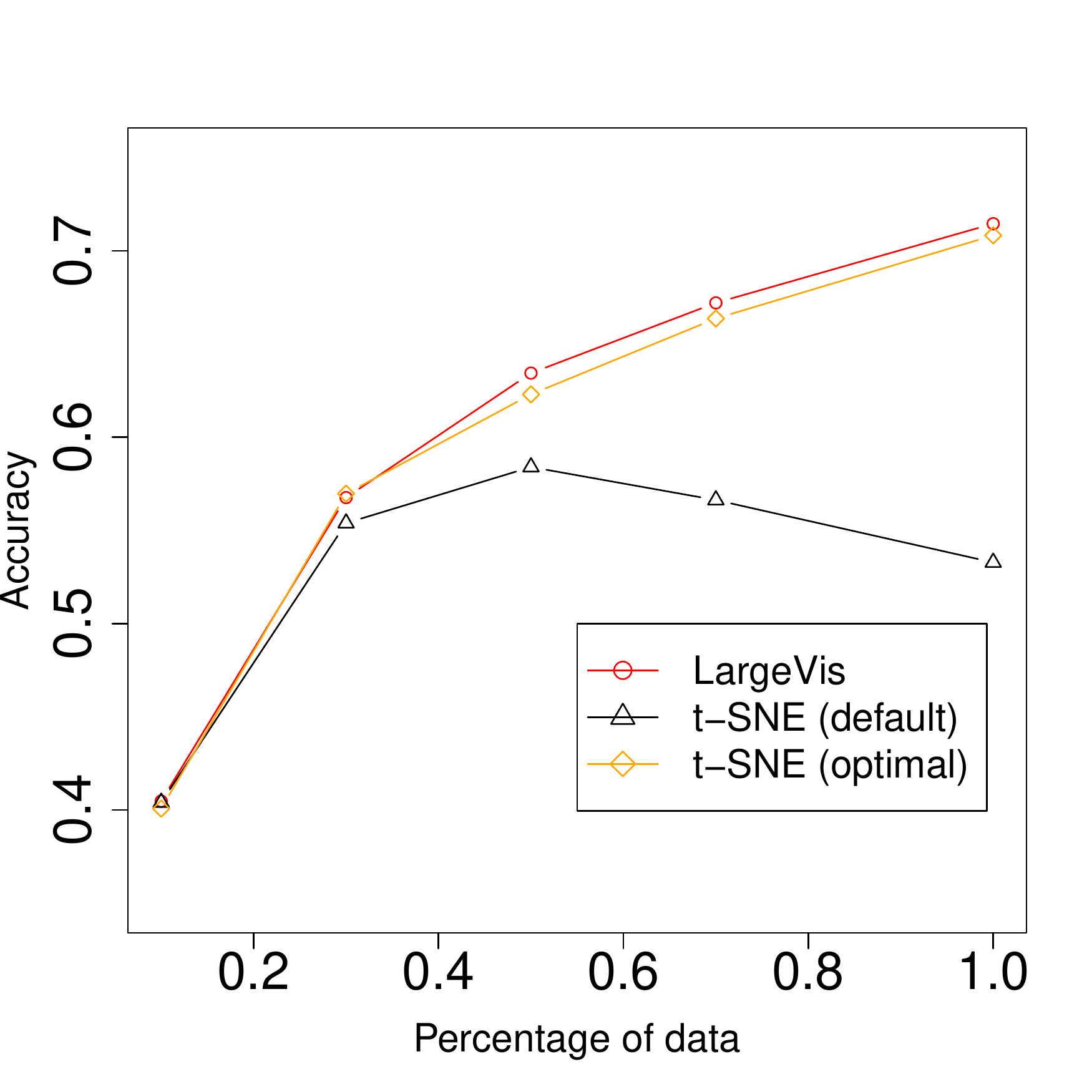}
	}
	\subfigure[Time (WikiDoc)]{
		\label{fig::time_vs_size_ww}
		\includegraphics[width=0.22\textwidth]{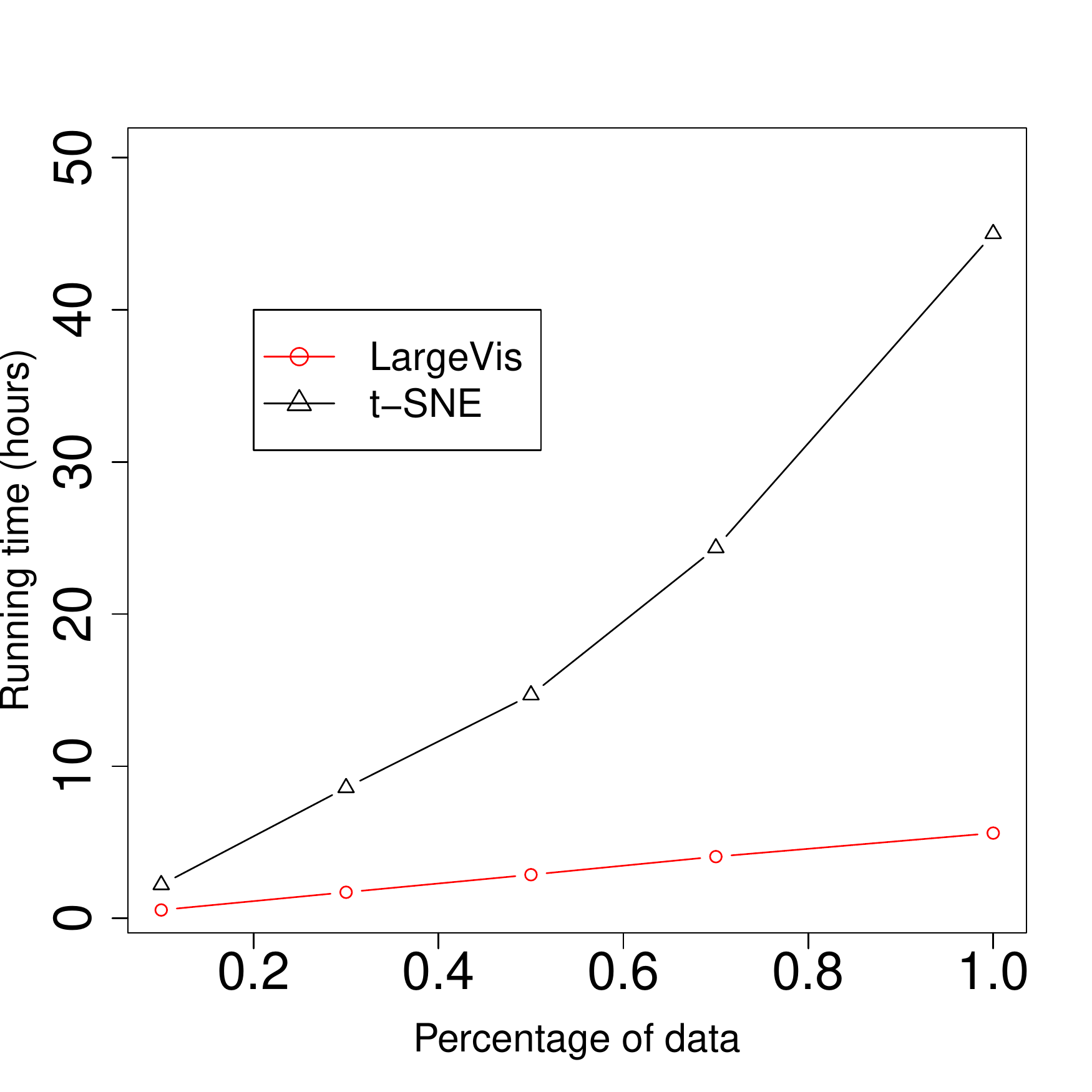}
	}
	\subfigure[Time (LiveJournal)]{
		\label{fig::time_vs_size_lj}
		\includegraphics[width=0.22\textwidth]{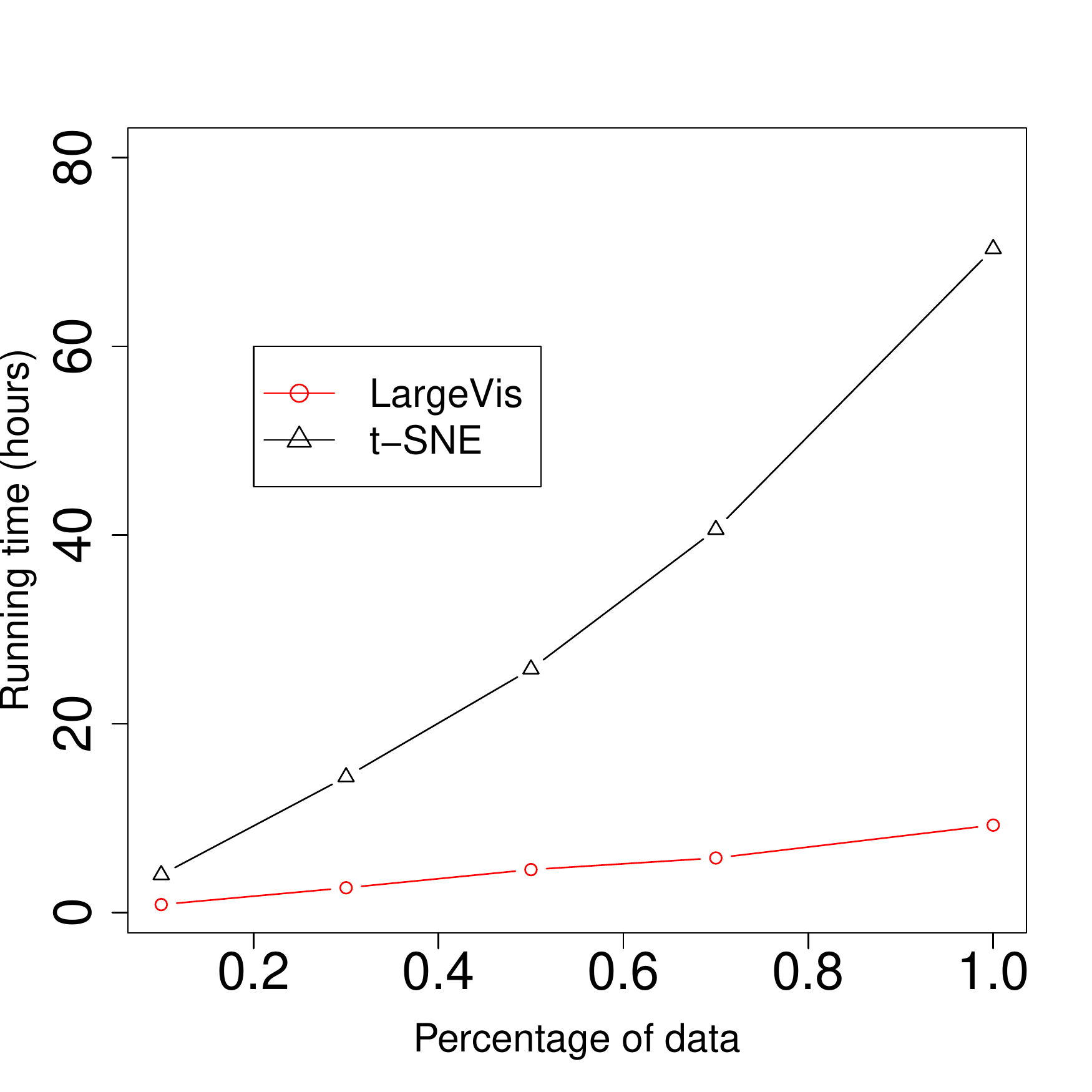}
	}	
	\caption{Accuracy and running time of the LargeVis and t-SNE w.r.t the size of data. LargeVis is much more efficient than t-SNE when the size of the data grows.}
	\label{fig::results_vs_size}
\end{figure*}

We further compare the performance of the LargeVis and t-SNE w.r.t the size of the data in terms of both effectiveness and efficiency. Fig.~\ref{fig::results_vs_size} presents the results on the WikiDoc and LiveJournal data sets. Different sizes of data sets are obtained by randomly sampling different percentages of the data.  In Fig.~\ref{fig::accuracy_vs_size_ww} and \ref{fig::accuracy_vs_size_lj}, we can see that as the size of the data increases, by using the default learning rates, the performance of the LargeVis increases while the performance of t-SNE decreases. By exhaustively tuning the learning rates, the performance of t-SNE will be comparable to LargeVis. However, this process is very time consuming, especially on large-scale data sets.  Fig.~\ref{fig::time_vs_size_ww} and \ref{fig::time_vs_size_lj} show that the LargeVis becomes more and more efficient than t-SNE as the size of the data grows. This is because the time complexity of graph visualization in t-SNE is $O(N\log(N))$ while that of LargeVis is $O(N)$. 

\subsubsection{Parameter Sensitivity}

\begin{figure}[htdb!]
	\centering
	\subfigure[Negative samples]{
		\label{fig::ps_ns}
		\includegraphics[width=0.22\textwidth]{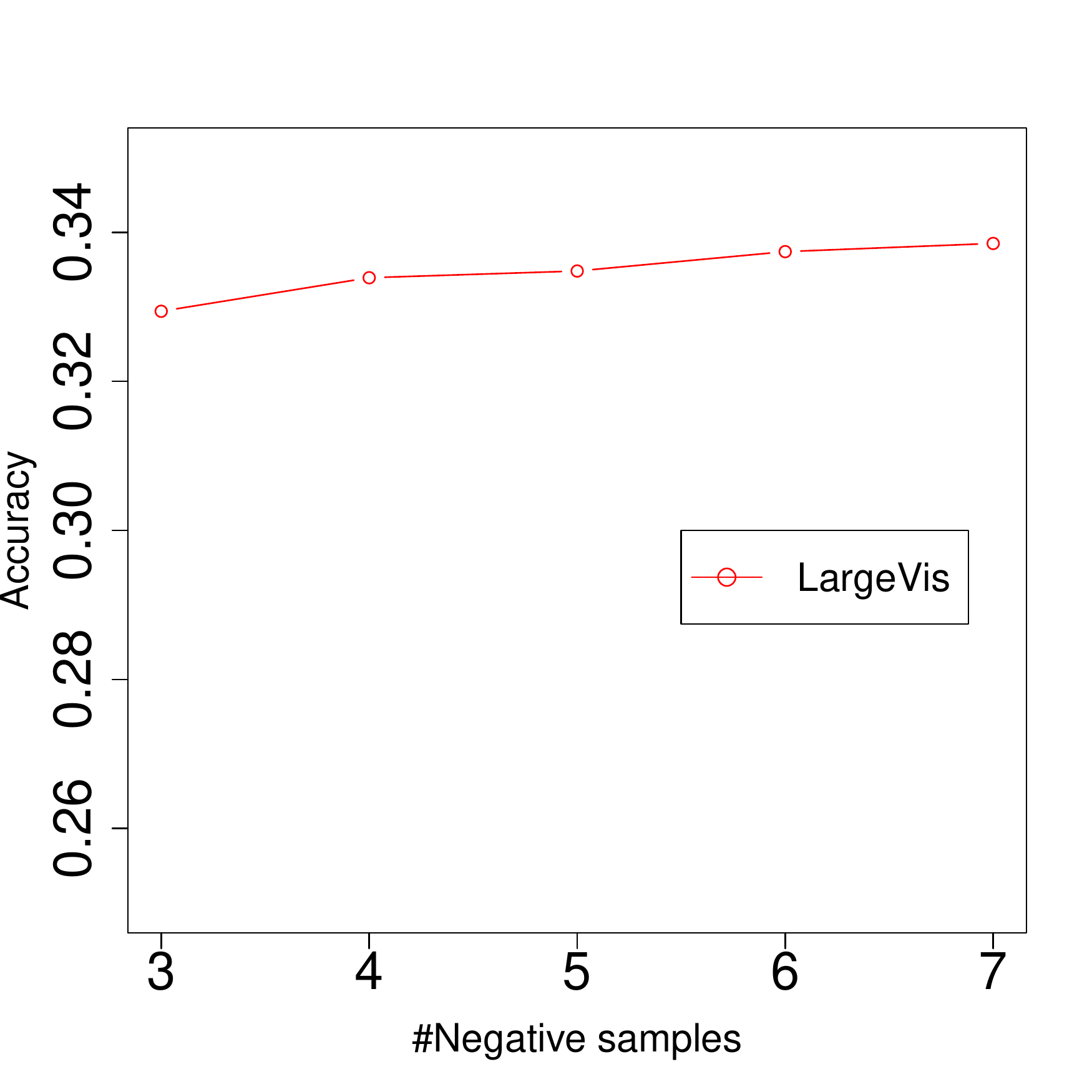}
	}
	\subfigure[Training samples]{
		\label{fig::ps_samples}
		\includegraphics[width=0.22\textwidth]{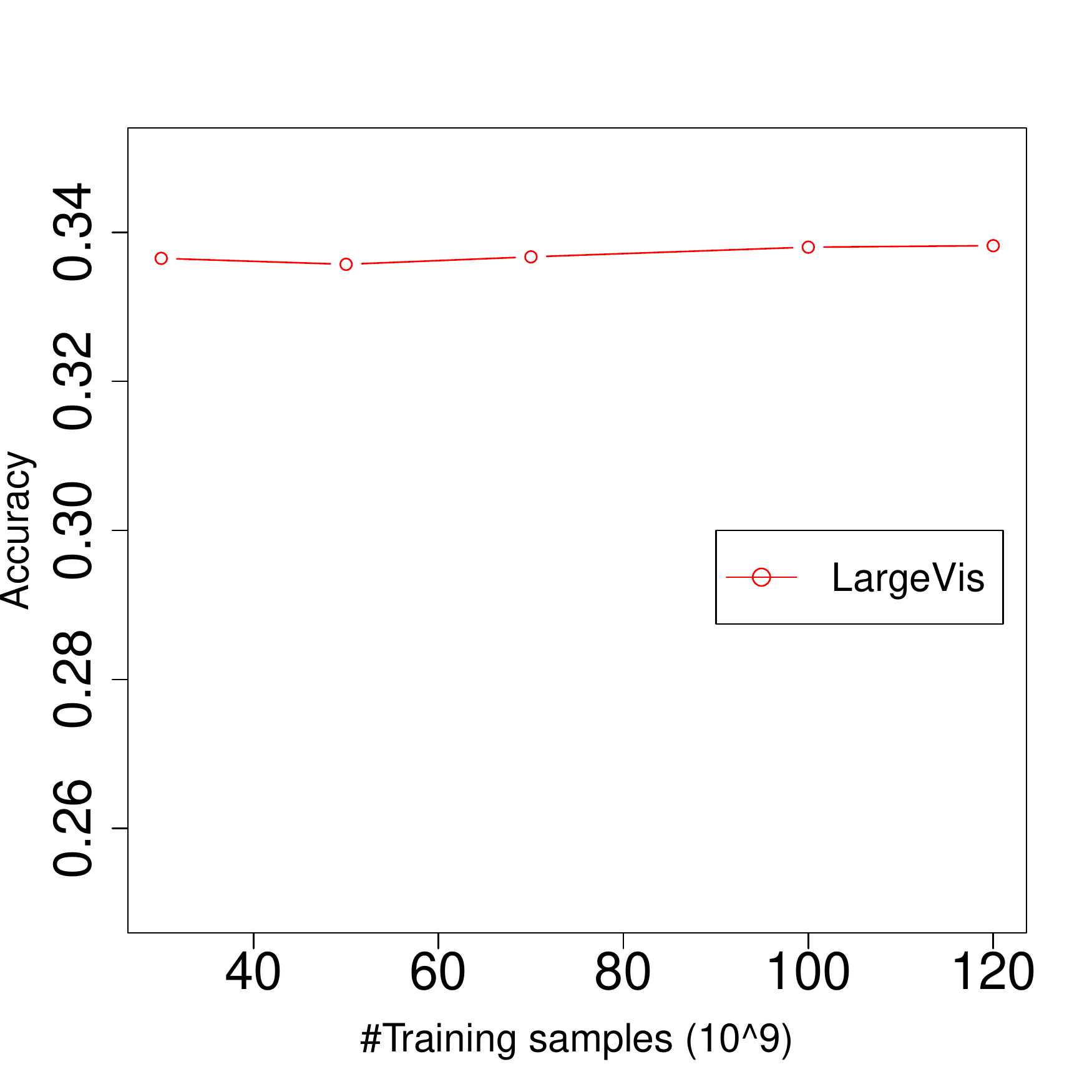}
	}	
	\caption{Performance of LargeVis w.r.t the number of negative samples and training samples on WikiDoc. Performance is not sensitive to the two parameters.}
\end{figure}

Finally, we investigate the sensitivity of the parameters in the LargeVis including the number of negative samples (M) and training samples (T). Fig.~\ref{fig::ps_ns} shows the results w.r.t the number of negative samples. When the number of negative samples becomes large enough (e.g., 5), the performance becomes very stable. For each data point, instead of using all the negative edges, we just need to sample a few negative edges according to a noisy distribution. An interesting future direction is to design a more effective noisy distribution for negative edge sampling. Fig.~\ref{fig::ps_samples} presents the results w.r.t the number of training samples. When the number samples becomes large enough, the performance becomes very stable. 

\begin{figure*}
	\centering
	\subfigure[20NG (t-SNE)]{
		\includegraphics[width=0.4\textwidth]{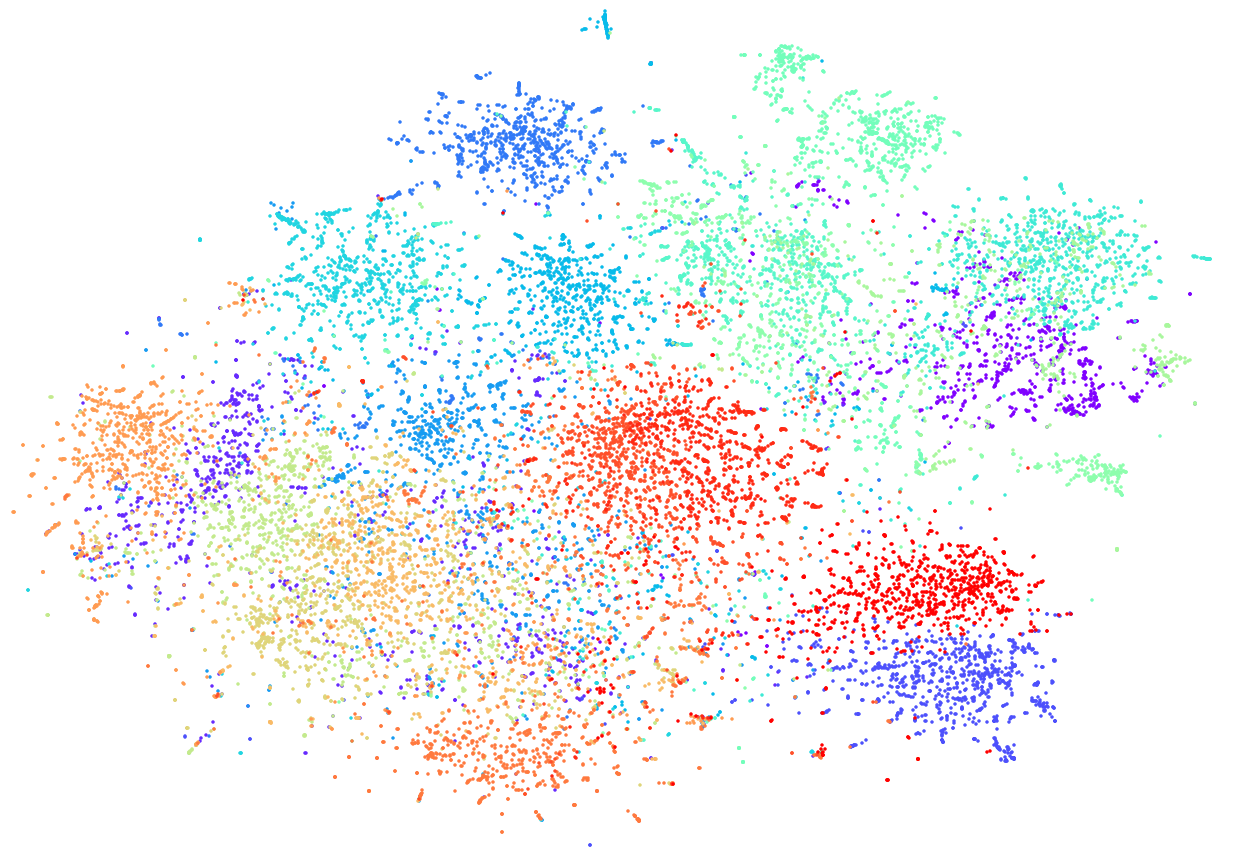}
	}
	\subfigure[20NG (LargeVis)]{
		\includegraphics[width=0.4\textwidth]{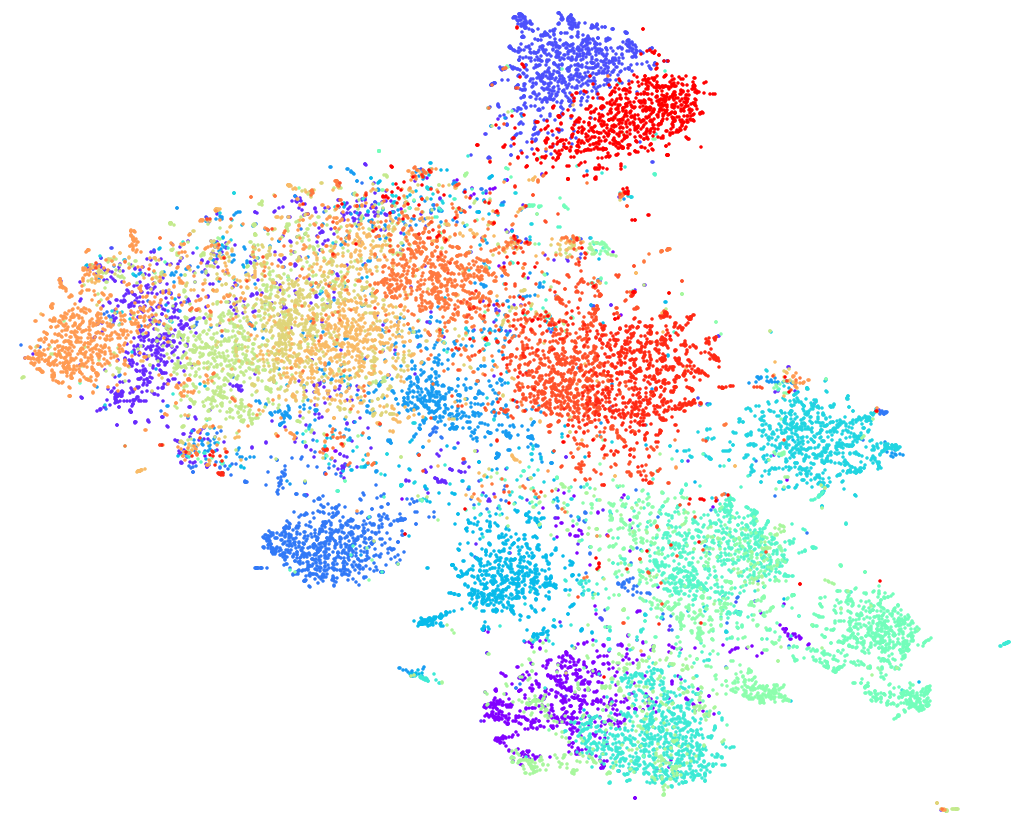}
	}	
	\subfigure[WikiDoc (t-SNE)]{
		\includegraphics[width=0.4\textwidth]{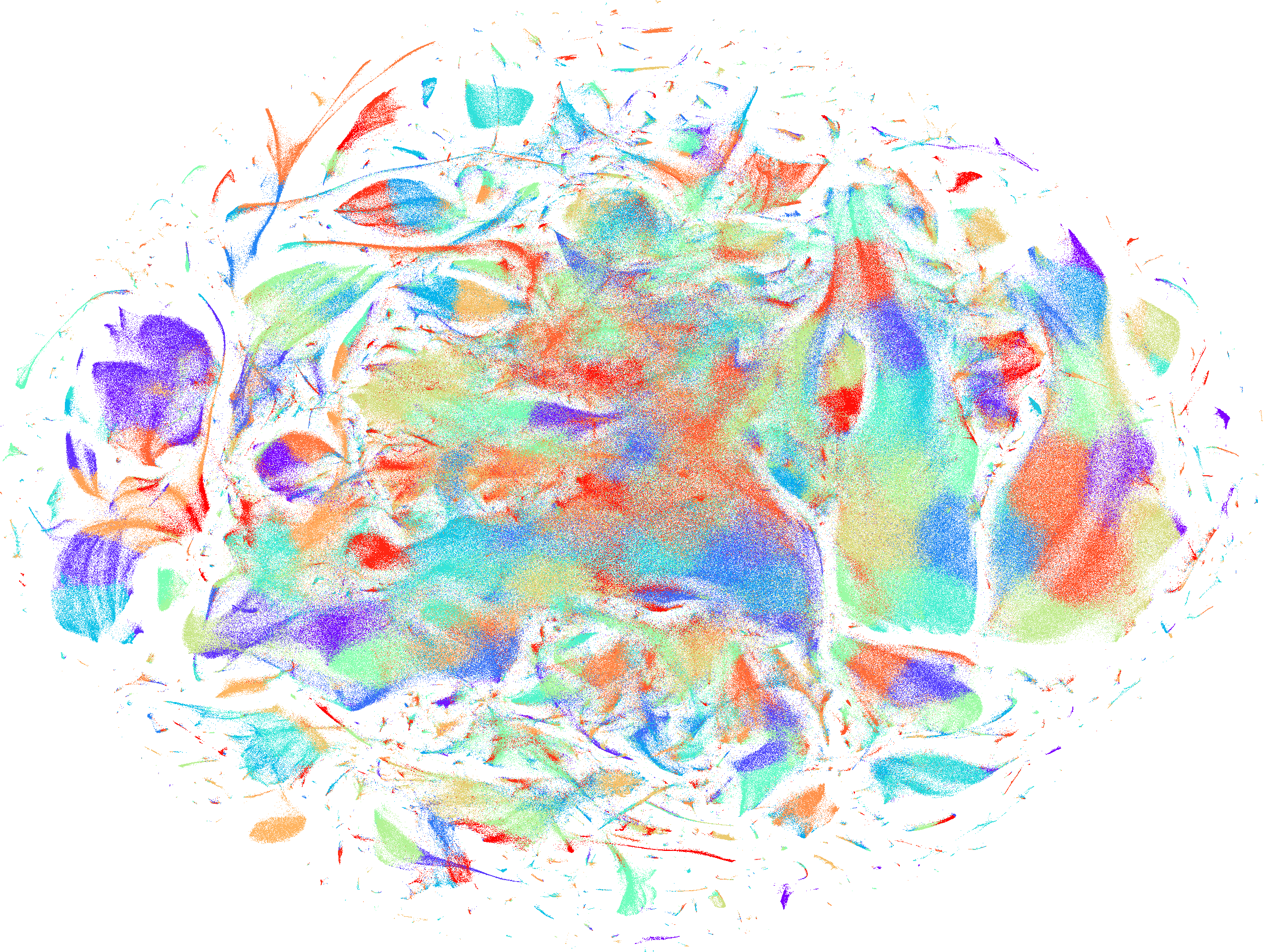}
	}
	\subfigure[WikiDoc (LargeVis)]{
		\includegraphics[width=0.4\textwidth]{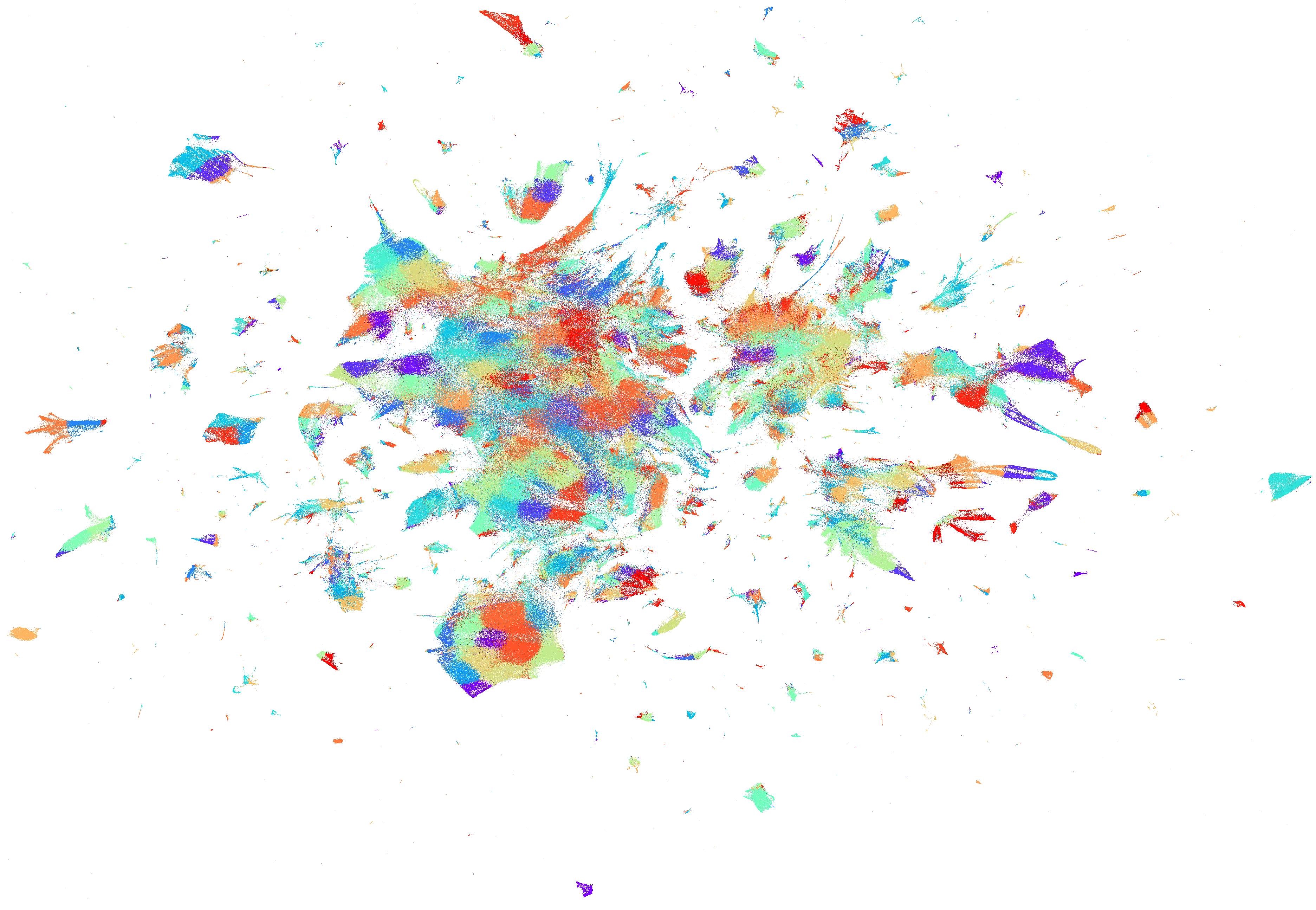}
	}	
	\subfigure[LiveJournal (t-SNE)]{
		\includegraphics[width=0.35\textwidth]{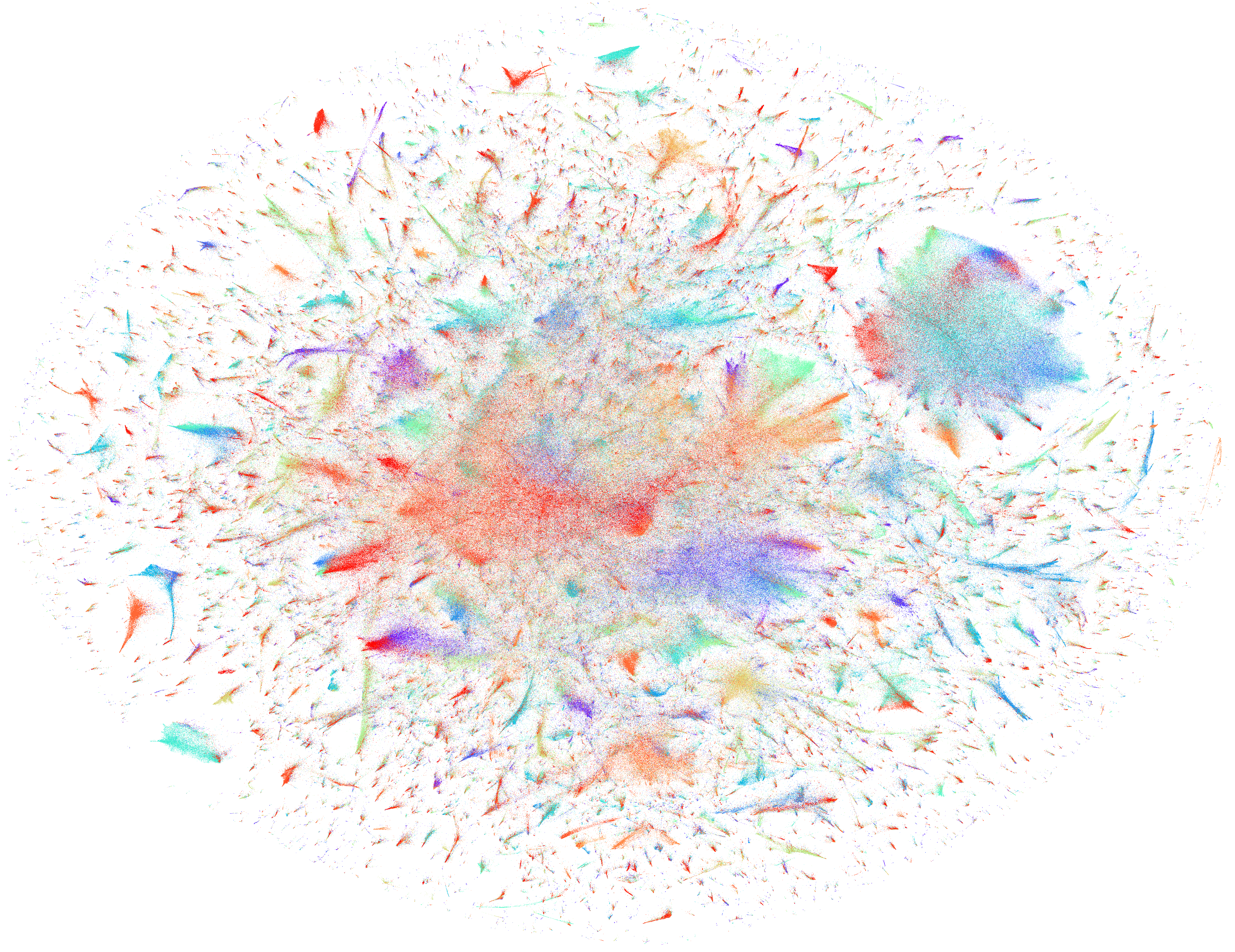}
	}
	\subfigure[LiveJournal (LargeVis)]{
		\includegraphics[width=0.35\textwidth]{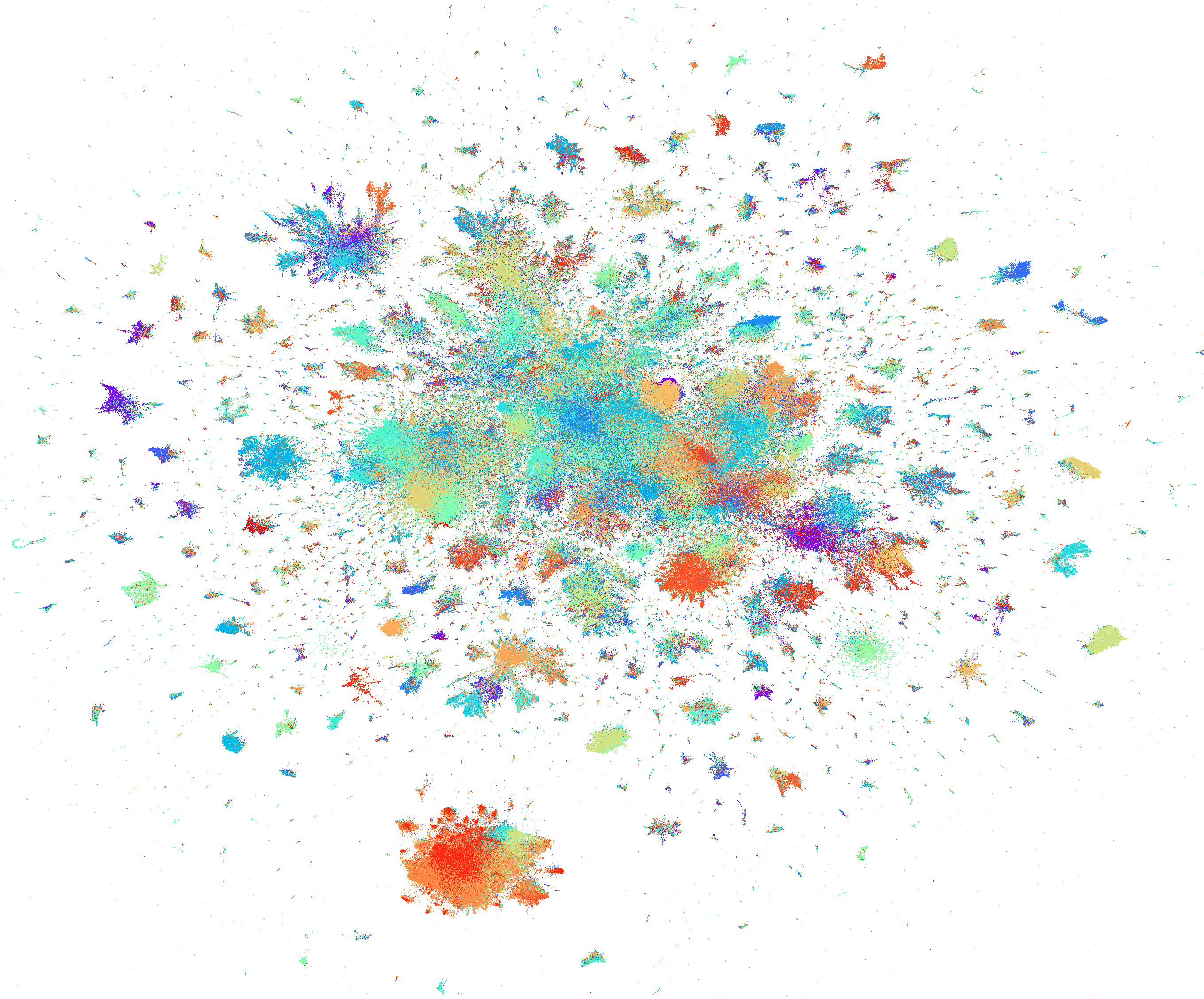}
	}	
	
	\caption{Visualizations of 20NG, WikiDoc, and LiveJournal by t-SNE and LargeVis. Different colors correspond to different categories (20NG) or clusters learned by K-means according to high-dimensional representations.}
	\label{fig::vis_batch}
\end{figure*}

\begin{figure*}
	\centering
	\subfigure[WikiWord (LargeVis)]{
		\includegraphics[width=0.35\textwidth]{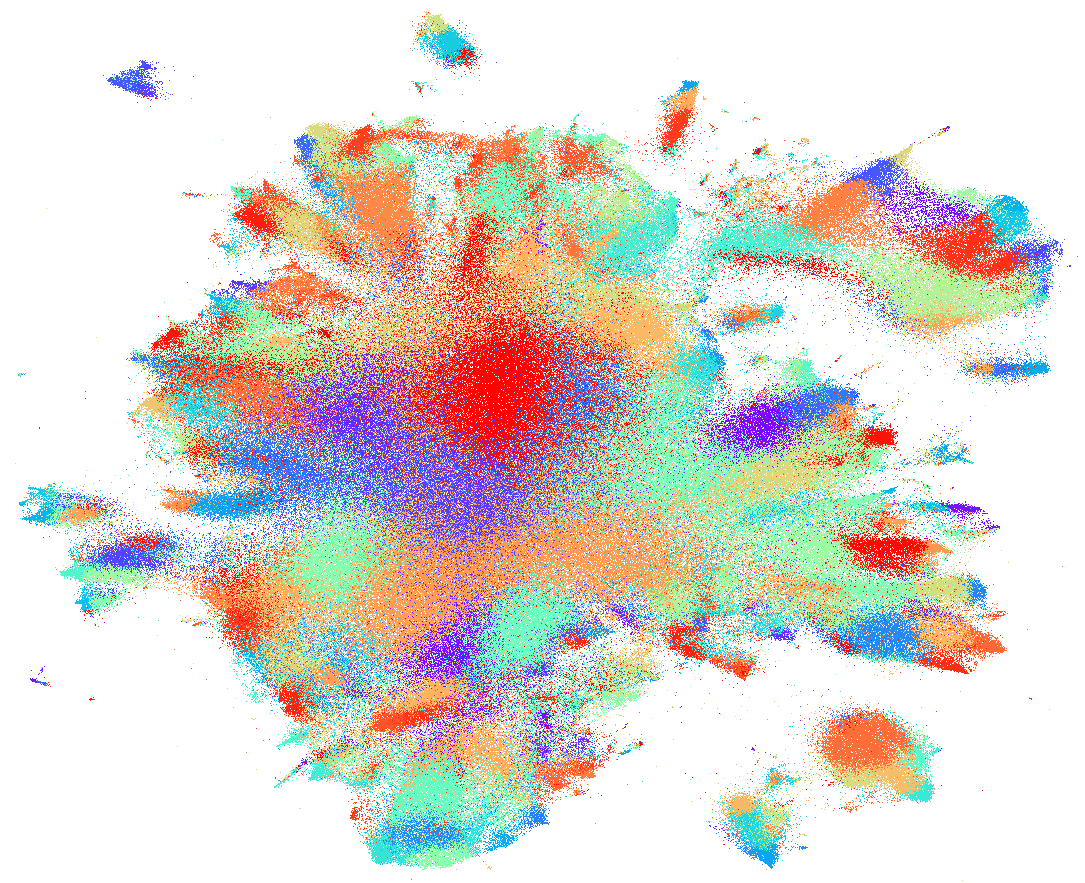}
	}
	\subfigure[CSAuthor (LargeVis)]{
		\includegraphics[width=0.4\textwidth]{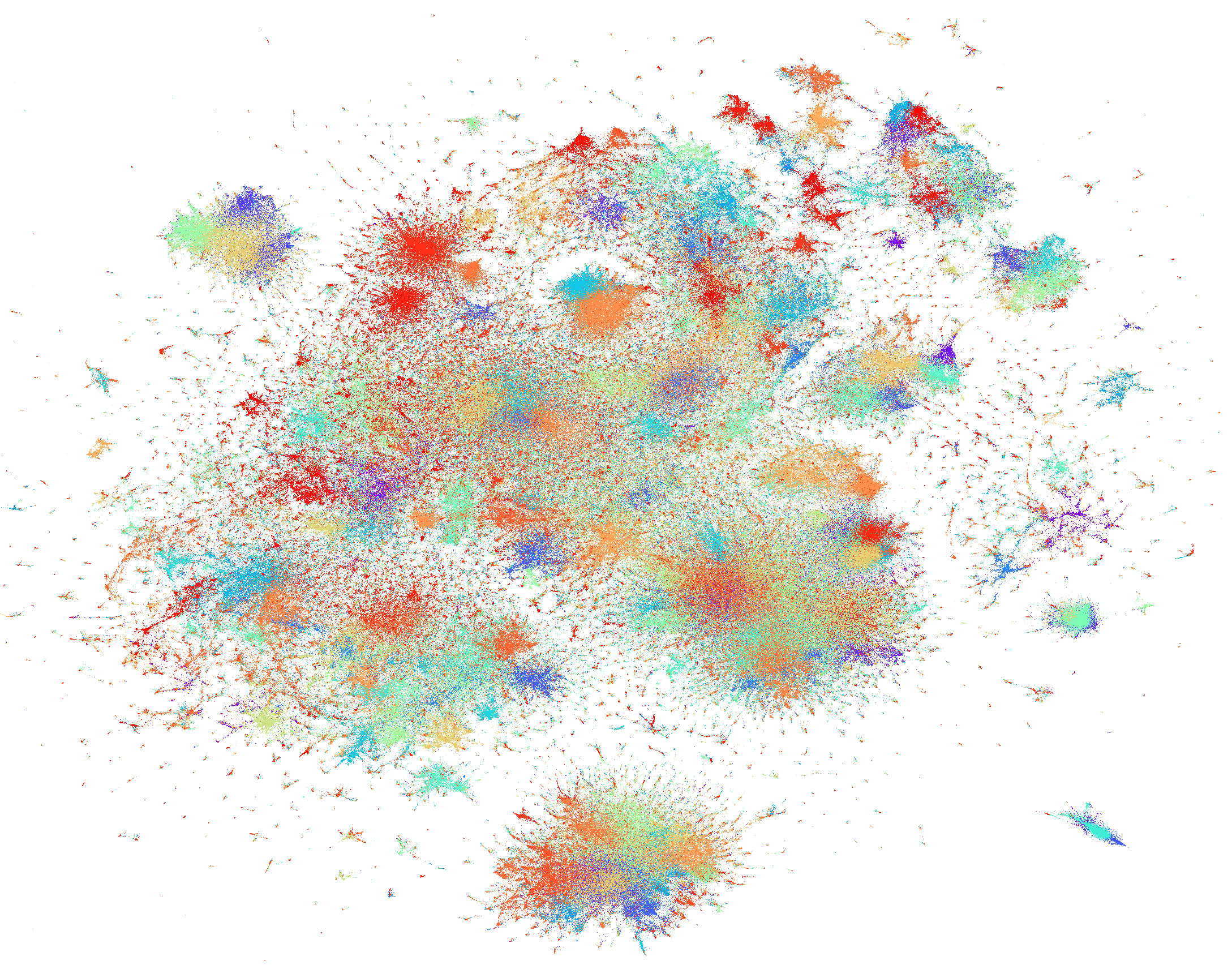}
	}
	\caption{Visualizations of WikiWord and CSAuthor by LargeVis. Colors correspond to clusters learned by K-means according to high-dimensional representations. }
	\label{fig::vis_batch_2}
\end{figure*}

\begin{figure*}[htdb!]
	\centering
	\includegraphics[width=0.8\textwidth, height=4in]{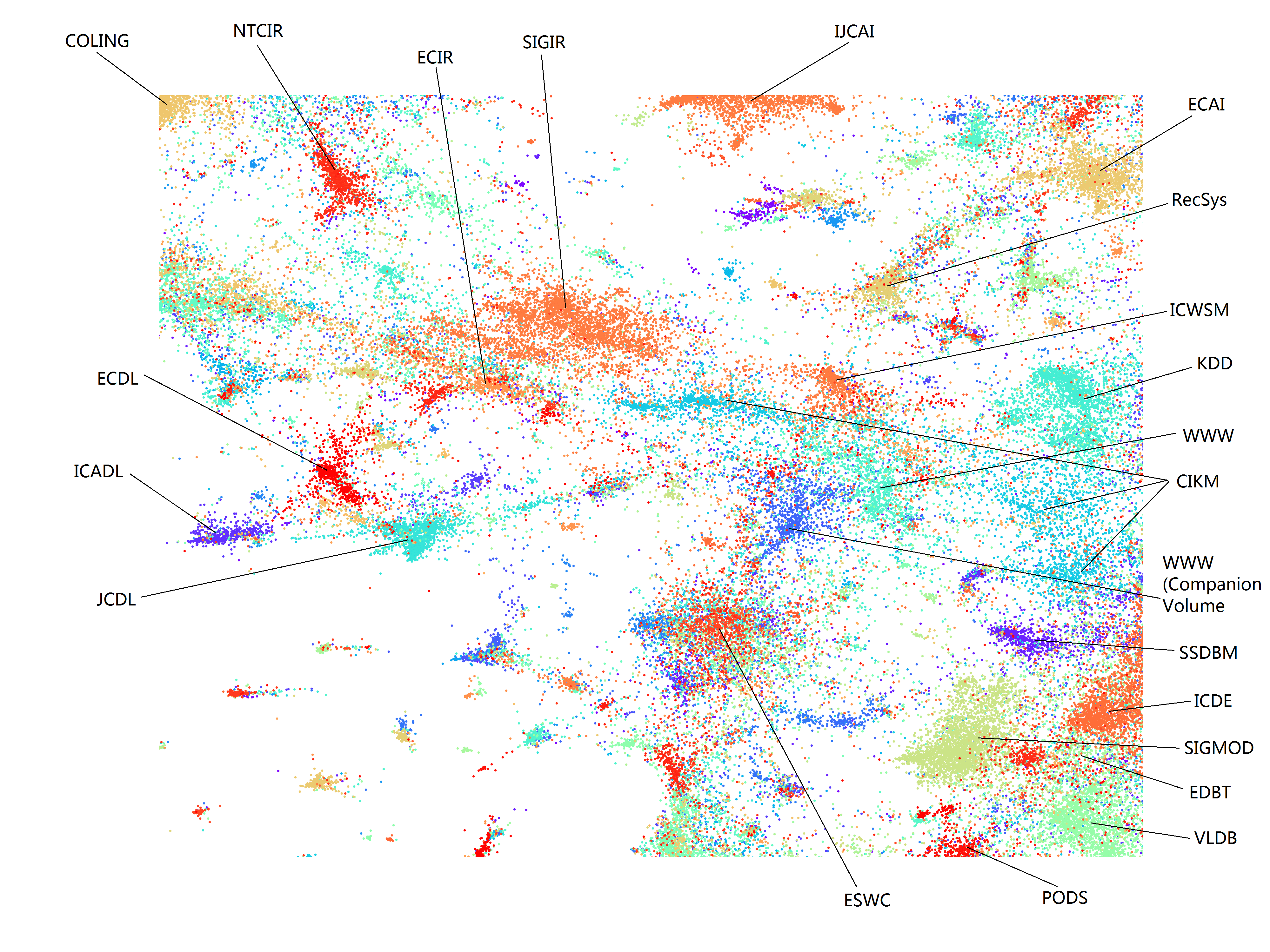}	
	\caption{Visualizing the papers in DBLP by LargeVis. Each color corresponds to a conference. }
	\label{fig::vis_dblp}
\end{figure*}

\subsection{Visualization Examples}
Finally, we show several visualization examples so that we can intuitively evaluate the quality of LargeVis visualizations and compare the performance of t-SNE and LargeVis. Fig.~\ref{fig::vis_batch} and Fig.~\ref{fig::vis_batch_2} present the visualizations. Different colors correspond to different categories (20NG), or clusters computed with K-means based on high-dimensional representations (WikiWord, WikiDoc, CSAuthors and LiveJournal). 200 clusters are used for all the four data sets. We can see that on the smallest data set 20NG, the visualizations generated by the t-SNE and LargeVis are both meaningful and comparable to each other. 
On the large data sets such as WikiDoc and LiveJournal, which contain at least 2.8 million data points, the visualizations generated by the LargeVis look much more intuitive than the ones by t-SNE.

Fig.~\ref{fig::vis_dblp} shows a region of the visualization of DBLP papers generated by LargeVis. Each color corresponds to a computer science conference. The visualization is very intuitive. The papers published at WWW are connected to the papers of ``WWW (Companion Volume),'' corresponding to its workshop and poster papers. The closest conference to WWW is ICWSM, right to the north. This ``Web'' cluster is close to SIGIR and ECIR on the west (the information retrieval community), with three digital library conferences close by. KDD papers locate to the east of WWW, and the database conferences ICDE, SIGMOD, EDBT and VLDB are clustered to the south of KDD. It is interesting to see that the papers published at CIKM are split into three different parts, one between SIGIR and WWW, and two between KDD and ICDE, respectively. This clearly reflects the three different tracks of the CIKM conference: information retrieval, knowledge management, and databases.

\section{Conclusion}
\label{sec::conclusion}
This paper presented a visualization technique called the LargeVis which lays out large-scale and high-dimensional data in a low-dimensional (2D or 3D) space. LargeVis easily scales up to millions of data points with hundreds of dimensions. It first constructs a K-nearest neighbor graph of the data points and then projects the graph into the low-dimensional space. We proposed a very efficient algorithm for constructing the approximate K-nearest neighbor graphs and a principled probabilistic model for graph visualization, the objective of which can be optimized effectively and efficiently. Experiments on real-world data sets show that the LargeVis significantly outperforms the t-SNE in both the graph construction and the graph visualization steps, in terms of both efficiency, effectiveness, and the quality of visualizations. In the future, we plan to use the low-dimensional layouts generated by the LargeVis as the basis for more advanced visualizations and generate many intuitive and meaning visualizations for high-dimensional data. Another interesting direction is to handle data dynamically changing over time.  

\vskip -1em
\section*{Acknowledgments}
The co-author Ming Zhang is supported by the National Natural Science Foundation of China (NSFC Grant No. 61472006 and 61272343); Qiaozhu Mei is supported by the National Science Foundation under grant numbers IIS-1054199 and CCF-1048168. 
 
 \clearpage
\bibliographystyle{abbrv}
\bibliography{sigproc}

\end{document}